# Complex Deep Learning Models for Denoising of Human Heart ECG signals

Corneliu T Arsene

School of Electrical and Electronics Engineering
University of Manchester, Manchester, UK
Email: arsenecorneliu@tutanota.de

*Abstract :* Effective and powerful methods for denoising real electrocardiogram (ECG) signals are important for wearable sensors and devices. Deep Learning (DL) models have been used extensively in image processing and other domains with great success but only very recently have been used in processing ECG signals. This paper presents several DL models namely Convolutional Neural Networks (CNNs), Long Short-Term Memory (LSTM), Restricted Boltzmann Machine (RBM) together with the more conventional filtering methods (low pass filtering, high pass filtering, Notch filtering) and the standard wavelet-based technique for denoising EEG signals. These methods are trained, tested and evaluated on different synthetic and real ECG datasets taken from the MIT PhysioNet database and for different simulation conditions (i.e. various lengths of the ECG signals, single or multiple records). The results show the CNN model is a performant model that can be used for off-line denoising ECG applications where it is satisfactory to train on a clean part of an ECG signal from an ECG record, and then to test on the same ECG signal, which would have some high level of noise added to it. However, for real-time applications or near-real time applications, this task becomes more cumbersome, as the clean part of an ECG signal is very probable to be very limited in size. Therefore the solution put forth in this work is to train a CNN model on 1 second ECG noisy artificial multiple heartbeat data (i.e. ECG at effort), which was generated in a first instance based on few sequences of real signal heartbeat ECG data (i.e. ECG at rest). Afterwards it would be possible to use the trained CNN model in real life situations to denoise the ECG signal. This corresponds also to reality, where usually the human is put at rest and the ECG is recorded and then the same human is asked to do some physical exercises and the ECG is recorded at effort. The quality of results is assessed visually but also by using the Root Mean Squared (RMS) and the Signal to Noise Ratio (SNR) measures. All CNN models used an NVIDIA TITAN V Graphical Processing Unit (GPU) with 12 GB RAM, which reduces drastically the computational times. Finally, as an element of novelty, the paper presents also a Design of Experiment (DoE) study which intends to determine the optimal structure of a CNN model, which type of study has not been seen in the literature before.

# 1. Introduction

Electrocardiography (ECG) is a widely accepted method in the medical cardiology domain for analysing of cardiac conditions of human patients [1]. However, ECG signals are often affected by noise, random or deterministic and artefacts. These errors mix with the ECG signal generated by the human heart, making it hard to extract underlying features and interpret the ECGs. The sources of errors can be due to various events such as movements of the patient, electromagnetic noise induction of the electronic hardware situated nearby, or muscular contraction artefacts.

A large number of methods to deal with noise and/or artefacts from ECG signals have been developed, such as adaptive Filters [2,3], Independent Component Analysis (ICA) [4], Empirical Mode Decomposition (EMD) [5], adaptive Fourier decomposition [6], Savitzky-Golay filter [7], threshold method for high frequency noise detection [8], Kalman filters [9], Bayesian filter framework [10], wavelet technique [11], clustering of morphological features [12], and Neural Networks [13]. Very recent attempts include arrhythmia heart classification using different Deep Learning (DL) models [14] and QRS characteristics identification using Support Vector Machines (SVM) [15]. These methods often do not take into account the problem of very high levels of noise present in the ECGs. Similarly, in [16] a first attempt to use DL based on the Long Short-Term Memory (LSTM) models for noise rejection in ECGs was proposed, while in [17] auto-encoders were investigated, but they did not consider drifting noise, which can be several times higher in magnitude than the ECG signal itself. This heavy and drifting noise is common in wearable sensors. Therefore, in this paper there are investigated several DL models for the removal and rejection of such noise in ECG signals. The ultimate scope is not necessarily to decrease the errors in the denoised ECG signals, but rather to compare different DL models on similar ECG real datasets even if some of the respective ECG datasets may not contain enough input features. However, the issue of minimization of the errors is also addressed, for example, by employing various measures of evaluation of the quality of the denoised ECG signals.

The paper is structured as follows: a number of synthetic and real ECG datasets and measures of evaluation are described in section 2. Section 3 includes the denoising of the synthetic and real ECG datasets described in section 2. Section 4 represents the main gist of this paper and includes the followings: Design of Experiment (DoE) to determine the optimal CNN model, DL model train and test on record 118 of MIT PhysioNet database (section 4.1), DL model train and test on record 105 of MIT PhysioNet (section 4.2), comparison of Convolutional Neural Network (CNN) model with Restricted Boltzmann Machine (RBM) (section 4.3), DL model train and test on 10 records of MIT PhysioNet database and also test on 1 different record (section 4.4), denoising of multiple heartbeat ECG signals (section 4.5). The PhysioNet database is a very large research resource available on-line and which contains a large number of complex physiological signal data repositories [18] including ECG signal data repositories. The Massachusetts Institute of Technology (MIT) in United States (US) and Boston's Beth Israel Hospital (BIH) completed and started to distribute an important ECG database (i.e. MIT-BIH Arrhythmia database) at the beginning of 1980s and which is available on-line [19].

## 2. *Examples of synthetic and real ECG data and measures of evaluation*

Initially in this study, there will be used briefly two synthetic ECG datasets so that to gain a good understanding on the topic of denoising of ECG signals.

Two datasets comprise synthetic ECG data generated with an ECG synthetic data generation software similar as in [20] (i.e. other similar and more sophisticated software [21, 22]), while a third dataset is a real dataset. Each dataset was divided in a training (3/4) and a testing dataset (1/4). The first synthetic dataset has 6888 clean ECG signals, each signal with a duration of 10 seconds and with 30000 samples per ECG signal (sampling rate 3000 Hz). The size of these synthetic datasets of 6888 ECG signals was determined by experiment that is by checking the errors in the DL model predictions over the testing dataset.

It is well-known that in order to train well a DL model, there has to be available a satisfactory number of training patterns, which to cover the variability of the input features. However, it is not always possible to do this and a solution is to generate artificially realistic training patterns. This solution will be used as well in this work. However, it will be of interest to train the DL models and to study their performances based also only on the realistic training patterns, which are available in various online repositories without generating any other additional realistic training patterns.

The duration of 10 seconds of an ECG signal was also chosen especially for the purposes of investigating the performances of the DL models. This is not the usual method for ECG signal processing, which signals are usually investigated on a second by second basis.

The ECG signal is varied between 57 to 67 heartbeats per minute as a test case scenario. The voltage varies between 1 mV to 3 mV. Fig. 1 shows an example of a 10 seconds normal synthetic ECG signal with multiple heartbeats. The P-wave is the contraction or depolarization of the human heart atria, QRS-complex is the contraction or depolarization of human heart ventricles, T-wave is the repolarization of human heart ventricles [1].

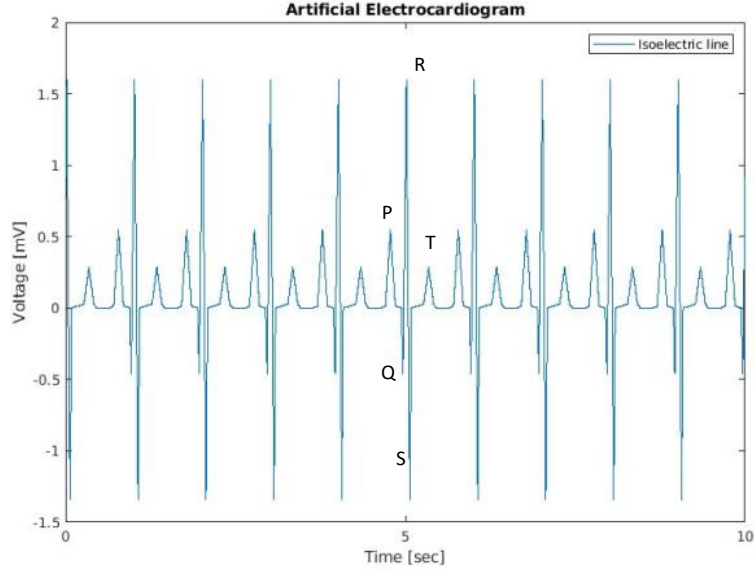

Figure 1. Synthetic normal ECG signal (i.e. duration 10 seconds) without noise comprising 30000 sample points and which is the desired output of the DL models.

The motion equations, which describe a single P-Q-R-S-T wave of the type shown in Fig. 1 are based on the dynamical model described in [23]:

$$\dot{x} = \alpha x - \omega y$$
$$\dot{y} = \alpha y + \omega x \qquad (1)$$
$$\dot{z} = -\sum_{i \in \{P,Q,R,S,T\}} \alpha_i \Delta\theta_i \exp(-\Delta\theta_i^2/2b_i^2) - (z - z_0)$$

where the model calculates a trajectory in a three-dimensional space with co-ordinates $(x, y, z)$, $\theta_P, \theta_Q, \theta_P, \theta_R, \theta_T$ are angles on a unit circle in $(x,y)$ dimensional space which encompasses one heartbeat cycle formed of the P, Q, R, S, T spikes, $z_0$ is a baseline value, ω is the angular velocity of the trajectory as it moves around the unit circle in $(x,y)$ plane and it is connected to the heartbeat rate as $2\pi f$, $f$ is frequency measured in Hertz, $\Delta\theta_i = (\theta - \theta_i) mod(2\pi)$, $\alpha = 1 - \sqrt{x^2 + y^2}$, $\theta = atan2(y, x)$.

In [23] the equations of motions were integrated numerically by using a fourth order Runge-Kutta method (i.e. fixed time step $\Delta t = 1/f$) and the $z$ variable from the system of eqs. (1) produces a synthetic ECG cycle with a realistic PQRST heartbeat cycle as shown in Fig. 1.

The first synthetic dataset also includes 6888 noisy ECG signals, which replicate the same noise conditions as found in real signals [24]. Specifically, the noise can be two or three times of magnitude of the ECG signal (Fig. 2(a)) (i.e. signal-to-noise ratio (SNR) = -3dB). A further strong drift is added to the random noise for only 861 signals of the 6888 noisy ECG signals, as shown in Fig. 2(b) (SNR= -7dB). The drift may correspond to various events such as sudden movement of the patient or random limb movement.

The second synthetic dataset has 6888 clean ECG signals and 6888 noisy ECG signals and all noisy signals have both random and strong drifting noise (SNR=-7dB) of various levels of drifts.

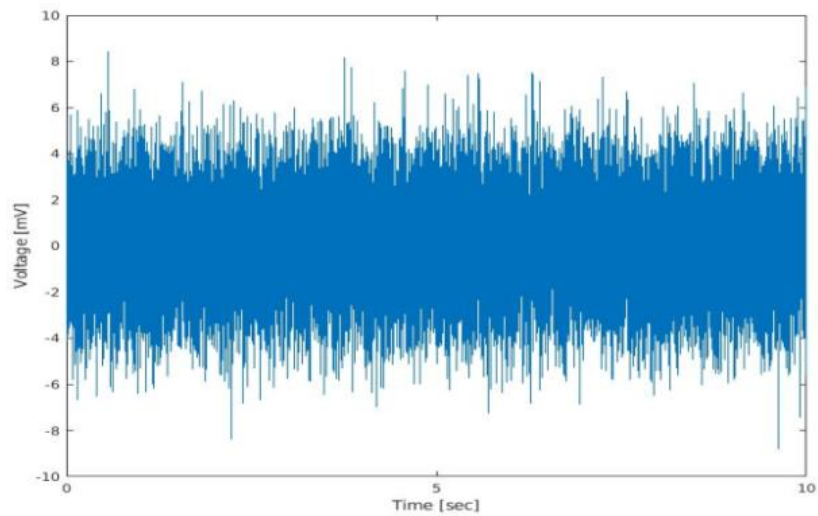

(a)

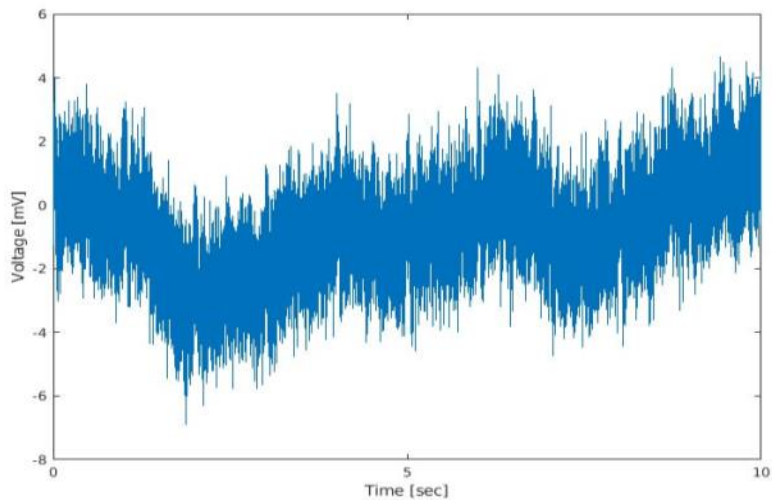

(b)

Figure 2. ECG signal affected by noise: a) ECG signal affected by two times random noise. b) ECG signal affected by strong drifting plus random noise.

The third dataset contains real ECG signals [25] that is record 118 of MIT-BIH Arrhythmia Database, which can be affected by various types of noise [26] (i.e. baseline wander, muscle artefact, electrode motion artefact). Each record was obtained from 2 channels at 360 samples per second with a total duration of 30 minutes. The dataset is in the PhysioNet WaveForm DataBase format (WFDB) [27]. The scope is to reject the electrode motion artefact from the ECG signal as the maximum of this type of noise could be four or five times or even higher than the normal ECG signal (Fig. 3).

In this section only, the real ECG signals from the above PhysioNet online repository was modified with artificially generated random and strong drifting noise (SNR=-7dB), while later in some of the following sections of this work, real noise will be added to the real ECG signals of this third dataset. For this real ECG dataset the training dataset has 5400 ECG clean real signals and another 5400 ECG noisy signals (i.e. noise artificially generated) with duration of 10 seconds. The testing dataset has 1800 ECG clean signals and another 1800 ECG noisy signals (i.e. noise artificially generated) with duration 10 seconds.

It is paramount to underline as already stressed in the neural network literature that a DL model needs enough data so that to train well and then to be possible to apply it successfully. For the third real ECG dataset from above, 180 different sequences of real ECG signals, each with duration of 10 seconds were used to generate the 5400 clean/noisy training signals and the 1800 clean/noisy testing ECG signals, each sequence with duration of 10 seconds. Later in this work, we would like to abstain if possible from doing this and eventually to use directly the ECG clean or noisy data which is available in various online repositories.

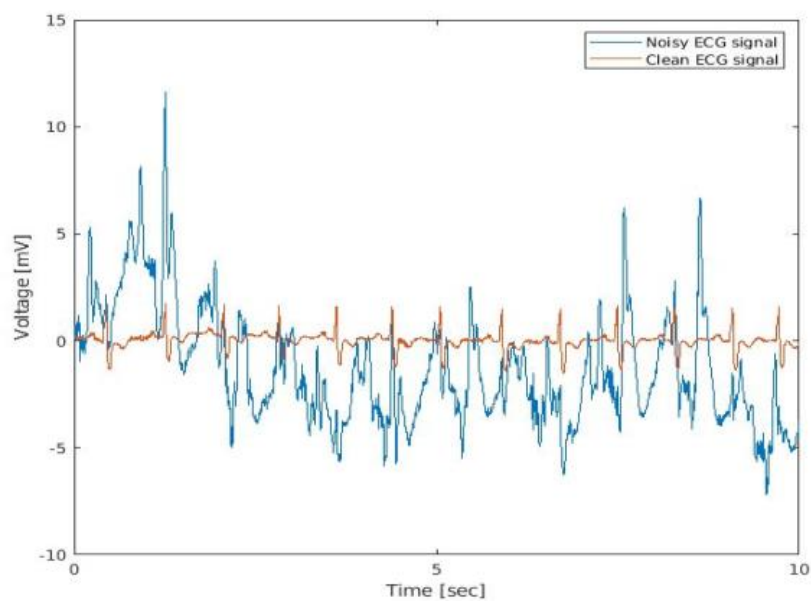

Figure 3. Real ECG signal affected by real electrode motion artefact.

In the following sections, the evaluation of the denoised ECG signals will be done not only visually but also by using one or both measures from below that is Root Mean Square error (RMS) over a testing dataset as well as the Signal to Noise Ratio (SNR) over a testing dataset. These two measures are calculated in order to see the difference between a clean and a denoised ECG signal and they are calculated as an average figure over an entire testing dataset:

$$RMS = \sqrt{\frac{1}{N}\sum_{n=0}^{N}[y_{pred}(n) - y_{clean}(n)]^2} \qquad (2)$$

$$SNR = 10 log_{10} \frac{\sum_{n=1}^{N}[y_{clean}(n)]^2}{\sum_{n=1}^{N}[y_{pred}(n) - y_{clean}(n)]^2} \qquad (3)$$

where $y_{clean}$ is the clean ECG signal, $y_{pred}$ is the predicted ECG signal, *n* is an index, *N* is the total number of sample points forming an ECG signal, which can have a duration of 1 second, 10 seconds, or any other time duration chosen by us.

The engineering literature is abundant about discussions on what is a good or a bad RMS value or a threshold value for SNR. Statistical significance has been usually taken at 5% [28] or even lower depending on the domain of study. In this work, it is proposed that the RMS limit value to be taken 5% of the difference between the maximum and the minimum values of the ECG signal being investigated. These maximum and minimum values for a clean ECG signal usually do not go above 3 mV and below 40μV: however because of the different ECG apparatus/machines used to record the ECG signal, then the clean ECG signal may vary between 3 mV and -3 mV, which give a difference of 6 mV. This interval (i.e. [-3 3]) is also characteristic for the clean ECG signals used in this work. It results in an RMS limit value of 0.3 mV (i.e. 5% of 6mV), which will be used through the remaining part of this investigation.

With regard to an SNR threshold, values higher than 10 dB are usually considered acceptable, while others are reporting SNR higher than 6 dB also worth to be taken into consideration. This depends also on the type of signal to be investigated that is for example signals from DSL cables or wireless signals and so on. In this work it is of interest the denoising of real ECG signals with SNR much below 0 (e.g. SNR= -6 dB, SNR = -8 dB) and especially the identification of the spikes corresponding to the R points in the P–Q-R-S-T wave. Therefore it will be considered an SNR= 8 dB the threshold point above which it is considered that the denoised ECG signal may still carry some valuable information.

In Fig. 4 is shown a comparison between a clean ECG signal (i.e. duration 22 seconds) and noisy ECG signal with SNR=8 dB. It can be noticed in Fig. 4(a) that it is possible to identify visually, eventually all the spikes corresponding to the R points of the ECG signal that is 30 spikes in total.

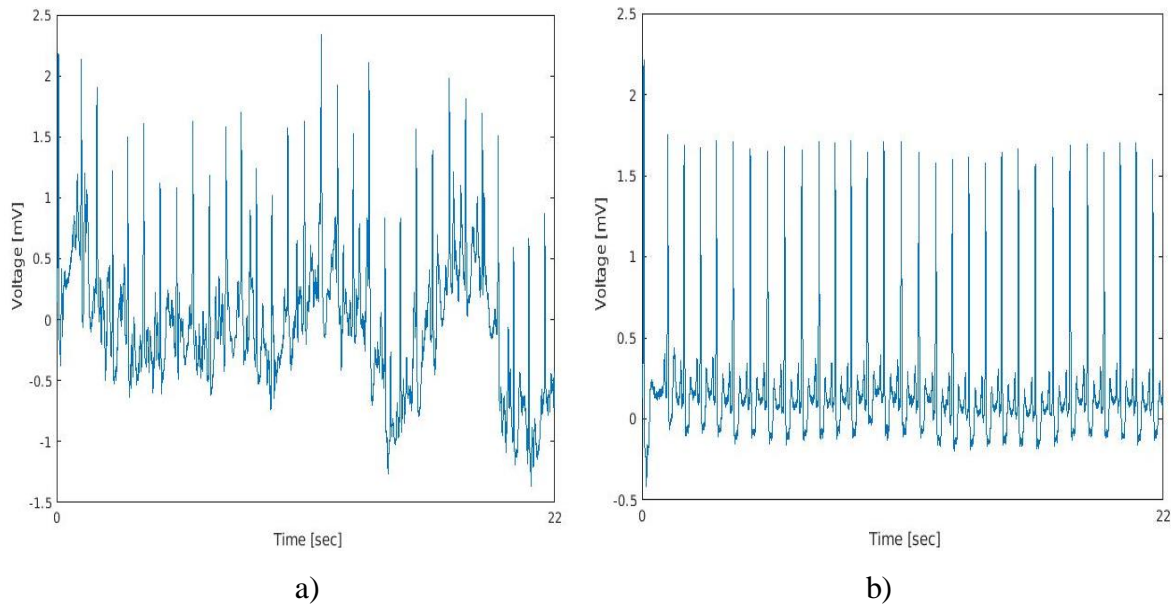

Figure 4. Comparison between a clean ECG signal and a typical noisy ECG signal with SNR=8 dB: a) noisy ECG signal; b) clean ECG signal.

As already envisaged above, initially there will be used three datasets (i.e. 2 synthetic datasets, 1 real dataset) to which will be added some synthetic noise with a maximum SNR of -7 dB.

However, it is of interest to use real ECG signals affected by real noise not only synthetically generated: MIT-BIH Arrhythmia database [27] contains a number of 48 different ECG records. Onto these ECG records it can be added real noise of different types such as baseline wander, muscle artefact, electrode motion artefact. Each record was obtained from 2 channels at 360 samples per second with a total duration of 30 minutes.

The real ECG noise can be added to these records by using the software function *nst.c* written in the C programming language and available from the PhysioNet web site [28]. The interest in this work is in the denoising of the ECG signals affected by the highest level of noise given by the electrode motion artefact: the way *nst.c* works is by adding the real noise, characteristics for example to the electrode motion artefact, to the clean ECG signal. It is intended to use in this study, in the context of the noisy ECG signals, patterns generated especially by the application *nst.c*. This may produce a limited number of training patterns, which in turn may affect the accuracy of the numerical results obtained with the various DL models. However, this is of interest herein, for example, which is the DL model, which is producing the most accurate results given an input ECG dataset, which may not be always well represented.

Other interest is in the ECG data itself, whether given an ECG real dataset what is the expectancy to denoise well the ECG signal and what other options would be available: for example generation of additional artificial input features based on the existent ones and which can be used for training.

Therefore, as already envisaged above, the scope is to reject the electrode motion artefact from the ECG signals as the maximum of this type of noise could be four or five times or even higher than the normal ECG signal.

Finally, the DL models, which will be developed and evaluated in the following sections, will be left to run for various periods of time so that to investigate, for example, whether any gains (e.g. accuracy of results) can be obtained from training a DL model for longer than normal computational times. Normal computational time can be regarded as the reference computational point time beyond which there is not noticed any further clear decrease in the error calculated (e.g. loss function) by the training algorithm used by the DL model. However, there can still be noticed a further decreased in the error calculated for example by other means such as over a testing dataset chosen by the investigator.

## 3. *Synthetic and Real ECG data denoising*

It is of interest to investigate from the multitude of DL models, which one would be the most suitable for ECG denoising from the point of view of accuracy of results. In this section there are presented results from applying the DL models on the previous described three datasets (i.e. two synthetic datasets, one real dataset).

Each of the three datasets was divided in a training (3/4) and a testing dataset (1/4). Each of the two synthetic datasets has 6888 clean ECG signals, each signal with a duration of 10 seconds and with 30000 samples per ECG signal (sampling rate 3000 Hz). As already described in the previous section, the difference between the two synthetic datasets is in the number of input training features, which have strong drifting (i.e. SNR=-7). The first synthetic dataset has less input training features with strong drifting than the second synthetic dataset. The third dataset is a real ECG dataset on which it was added synthetic generated noise with SNR equals -7.

The first DL model investigated is based on the Convolutional Neural Networks (CNNs) [29] implemented in Matlab [30]. Matlab gives the possibility of using kernels of type 9x9 or 23x23 for one dimensional ECG signal with duration of 1 second and, for example, of sampling size [360x1]. The first value 1.5 mV of the ECG signal in Fig.5 is used and a padding is created by the CNN DL toolbox of Matlab around the respective value, which has dimensions of 4, 4, 4, and 4 for above, below, right and left locations. This corresponds to a kernel size 9x9. There are other possible scenarios that worth to be investigated for kernels such as kernels with sizes 9x1 or 23x1 or even higher 71x1, and this will be also discussed. It is important also to investigate the number of filters in a convolutional layer or the number of convolutional layers to be used by a CNN model.

In Fig. 6 an ECG signal is investigated with a filter (i.e. CNN model) with kernel size 19x1 (i.e. experimentally determined).

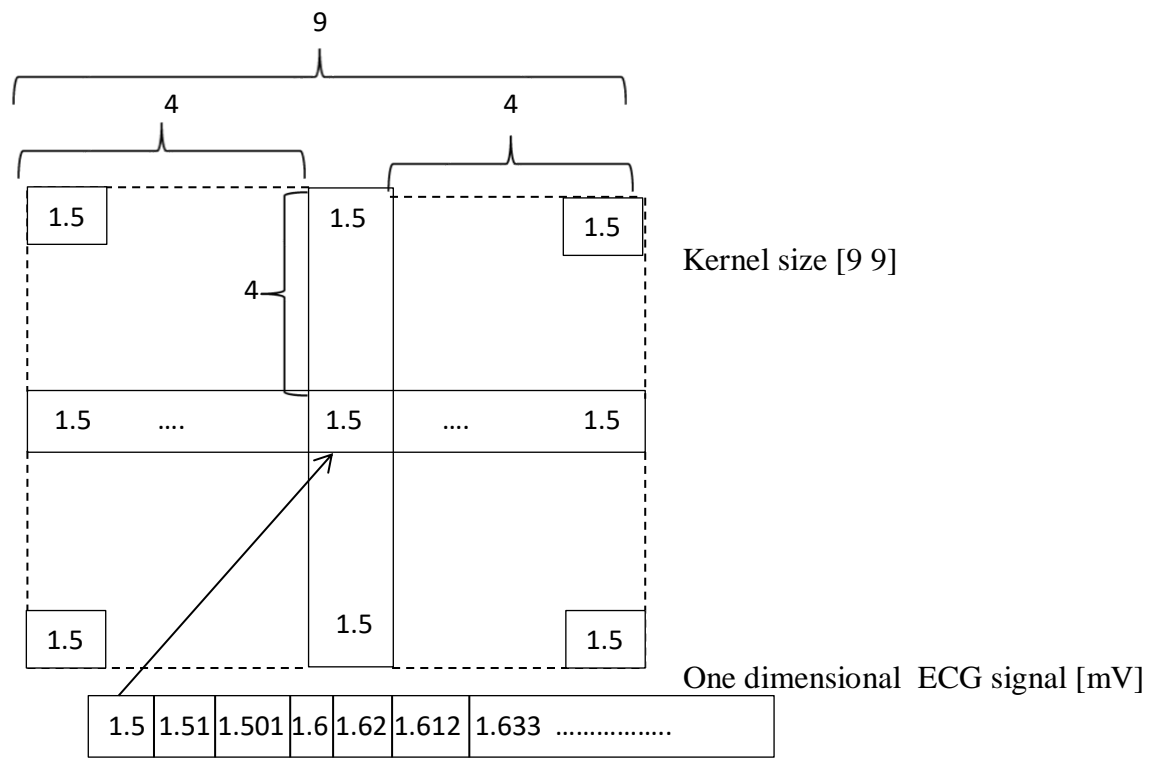

Figure 5. ECG signal investigated with kernel size 9x9 for a filter of a CNN model.

The CNN models have been used before for noise detection in ECG [31] but not yet for ECG reconstruction.

The CNN model used in this section was obtained by experiment and it consists of six 2-dimensional convolutional layers, each having 36 filters with kernel size of 19x1 per filter. The first layer is an input layer of size 30000x1x1 with 'zerocenter' normalization. $M = 30000$ is the number of samples per input ECG signal sequence of 10 seconds. Each convolutional layer has neurons that connect to parts of the input feature or connect to the outputs of the previous layer. The step size (i.e. stride) for the kernels is [1 1] while the padding is introduced so that the output is the same size as the input. Each convolutional layer is followed by a batch normalization layer with 36 channels, a rectified linear unit (ReLU) layer and an average pooling layer with stride of 4 and pooling size of [1 1] (i.e. sub-sampling). The succession of these layers reduces the dimensionality of the input ECG signal sequences. Before the final regression output layer, the signal goes through a fully connected layer for regression. Table 1 details all 27 layers of the CNN model with their characteristics. Fig. 6(a) depicts the structure of the CNN model. It is possible to assume that the DL model will be able to learn suitable filters that can be used for noise reduction so that to enable the recovery of the original ECG signals. An epoch goes through the entire dataset while an iteration represents the calculation of the gradient and the network parameters for the mini-batch data.

The CNN model uses Adam optimizer and with a batch training data size of 300. It was also noticed that in this case the obtained CNN performances did not change much with increasing the batch sizes. As an example, the average RMS error calculated over an entire testing dataset was 0.0476 after 20 epochs. This took 5 minutes to run. However, for the scope of investigation, the CNN model was also left to iterate for a longer time consisting of 200 epochs, which took 53 minutes to reach an average RMS over the same testing dataset of 0.0278. The CNN model was implemented in a Matlab environment using the DL toolbox and an NVIDIA TITAN V GPU to shorten the computational times.

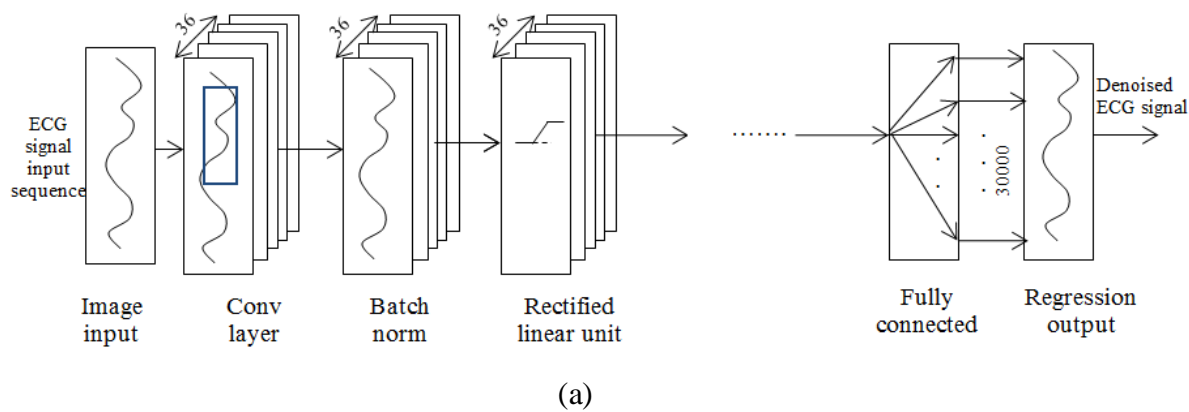

(a)

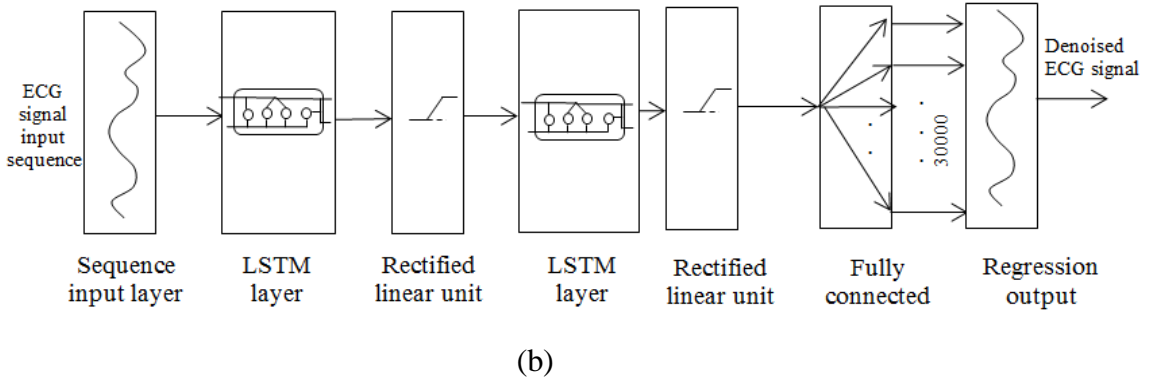

(b)

Figure 6. Structure of the DL models: a) CNN model. b) LSTM model.

The second DL model is based on the Long Short-Term Memory (LSTM) layer [32, 33] and it is also implemented in Matlab and using the DL toolbox. It consists of two LSTM layers with 140 hidden nodes per layer. The first layer is a sequence input layer with the dimension similar to the ECG sequence input signal of [30000x1]. The following layer is a LSTM layer with 140 hidden nodes empirically found. The LSTM layer has neurons that connect to the sequence input layer and also connect to the following layer, which is a rectified linear unit layer with 140 inputs. The second LSTM layer also has 140 hidden units and it is connected to the previous and subsequent rectified linear units. Similar to the CNN model, before the final regression output layer, the signal goes through a fully connected layer. More LSTM layers were found not to improve much the performance. Table 2 shows all 7 layers of the LSTM model and their properties, while Fig. 6(b) depicts the structure of the LSTM model. The model tries to learn long-term dependencies in the sequence input data although the CNN model tries to do the same thing by using kernels and eventually a deeper network structure. The LSTM model uses also the Adam optimizer for batch training and with a batch training data size of 300. A constant learning rate of 0.01 and a gradient threshold of 0.4 were used. For the same training dataset, the computational training time of the LSTM model was for 2000 epochs about 196 minutes and with an average RMS over the entire testing dataset of 0.2321. This value (i.e. 0.2321) is significantly higher than RMS=0.0476 obtained with the CNN model after only 20 epochs and 5 minutes run time (or 0.0278 after 200 epochs and 53 minutes run).

For comparison, the last model used here for noise rejection is the wavelet method, which has become popular for decomposing signals in many applications including ECG noise rejection. The wavelet transformation of an input signal $x(t)$ where $t$ is time is:

$$W_{a,b} = \int_{-\infty}^{\infty} x(t) \frac{1}{\sqrt{a}} \gamma^* \left(\frac{t-b}{a}\right) dt \qquad (4)$$

where $W_{a,b}$ is the wavelet transformation of $x(t)$, $a$ is the dilation parameter, $b$ is the location parameter, $\gamma^*(t)$ is the complex conjugate of the wavelet function which can be the Mexican-hat, Gaussian, or Daubechies wavelet function.

The wavelet method based on an empirical Bayesian method with a Cauchy prior and implemented in Matlab *wdenoise*() function [30] was used, with the default 'sym4', wavelet, where 4 is the number of vanishing moments. Other options of the function included denoising method as universal threshold and noise estimate as level independent. Various parameters and options were also tried.

TABLE 1. LISTING OF THE CNN LAYERS: $M = 30000$ IS THE NUMBER OF SAMPLES PER INPUT ECG SIGNAL WITH DURATION OF 10 SECONDS

| Nr | Type | Description |
|---|---|---|
| 1 | Image Input | 30000x1x1 images with 'zerocenter' normalization |
| 2 | Convolution | 36 19x1x1 convolutions with stride [1 1] and padding 'same' |
| 3 | Batch Normalization | Batch normalization with 36 channels |
| 4 | ReLu | Rectified Linear Unit |
| 5 | Average Pooling | 1x1 average pooling with stride [4 1] and padding [0 0 0 0] |
| 6 | Convolution | 36 19x1x36 convolutions with stride [1 1] and padding 'same' |
| 7 | Batch Normalization | Batch normalization with 36 channels |
| 8 | ReLu | Rectified Linear Unit |
| 9 | Average Pooling | 1x1 average pooling with stride [4 1] and padding [0 0 0 0] |
| 10 | Convolution | 36 19x1x36 convolutions with stride [1 1] and padding 'same' |
| 11 | Batch Normalization | Batch normalization with 36 channels |
| 12 | ReLu | Rectified Linear Unit |
| 13 | Average Pooling | 1x1 average pooling with stride [4 1] and padding [0 0 0 0] |
| 14 | Convolution | 36 19x1x36 convolutions with stride [1 1] and padding 'same' |
| 15 | Batch Normalization | Batch normalization with 36 channels |
| 16 | ReLu | Rectified Linear Unit |
| 17 | Average Pooling | 1x1 average pooling with stride [4 1] and padding [0 0 0 0] |
| 18 | Convolution | 36 19x1x36 convolutions with stride [1 1] and padding 'same' |
| 19 | Batch Normalization | Batch normalization with 36 channels |
| 20 | ReLu | Rectified Linear Unit |

| 21 | Average Pooling | 1x1 average pooling with stride [4 1] and padding [0 0 0 0] |
| 22 | Convolution | 36 19x1x36 convolutions with stride [1 1] and padding 'same' |
| 23 | Batch Normalization | Batch normalization with 36 channels |
| 24 | ReLu | Rectified Linear Unit |
| 25 | Average Pooling | 1x1 average pooling with stride [4 1] and padding [0 0 0 0] |
| 26 | Fully Connected | 30000 fully connected layer |
| 27 | Regression Output | Mean-squared-error with response 'Response' |

TABLE 2. LISTING OF THE LSTM LAYERS: $M = 30000$ IS THE NUMBER OF SAMPLES PER INPUT ECG SIGNAL WITH DURATION OF 10 SECONDS

| Nr | Type | Description |
|---|---|---|
| 1 | Sequenceinput: | Sequence input with 30000 dimensions |
| 2 | lstm_1: LSTM with 140 hidden units | InputWeights 560x30000; RecurrentWeights 560x140; Bias:560x1; HiddenState 140x1 |
| 3 | relu_1: ReLu | Rectified Linear Unit |
| 4 | Lstm_2: LSTM with 140 hidden units | InputWeights 560x140; RecurrentWeights 560x140; Bias 560x1; HiddenState 140x1 Cell 140x1 |
| 5 | relu_2: ReLu | Rectified Linear Unit |
| 6 | Fc: 30000 fully connected layer | Weights 30000x140; Bias:30000x1 |
| 7 | Regression output | Mean-squared error with response |

For the first synthetic dataset the predictions show that the CNN model is able to recover the original signal while the RMS value is 0.0198 for the result shown in Fig. 7(b).

For the second synthetic dataset, the LSTM predictions are less impressive with RMS value of 0.2201 in Fig. 8(b). The LSTM model was not able to train well on the first synthetic ECG dataset.

Over the first and second testing datasets the average RMSs calculated with the CNN model were 0.0278 and 0.0279, which is 8 times lower than 0.2321 the average RMS obtained over the second testing dataset with the LSTM model.

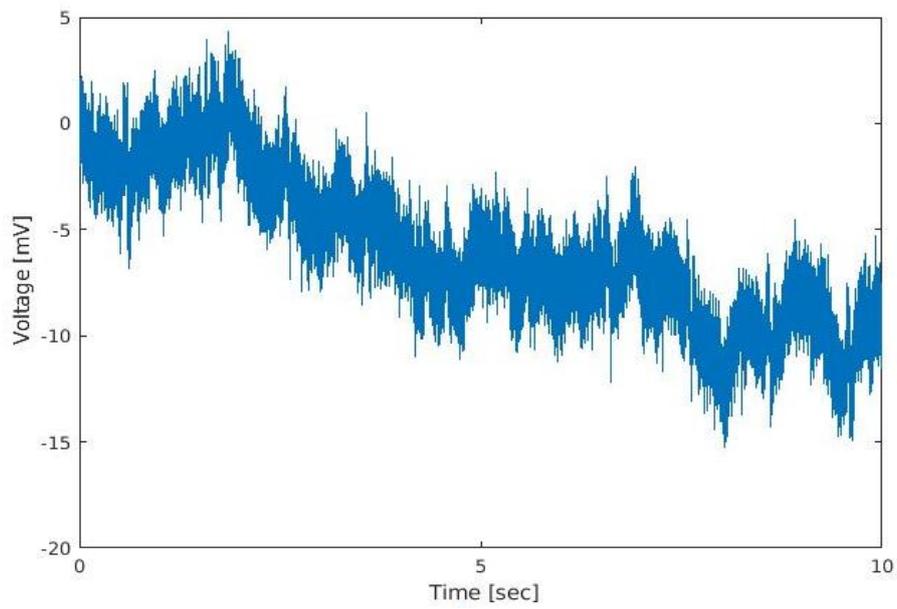

(a)

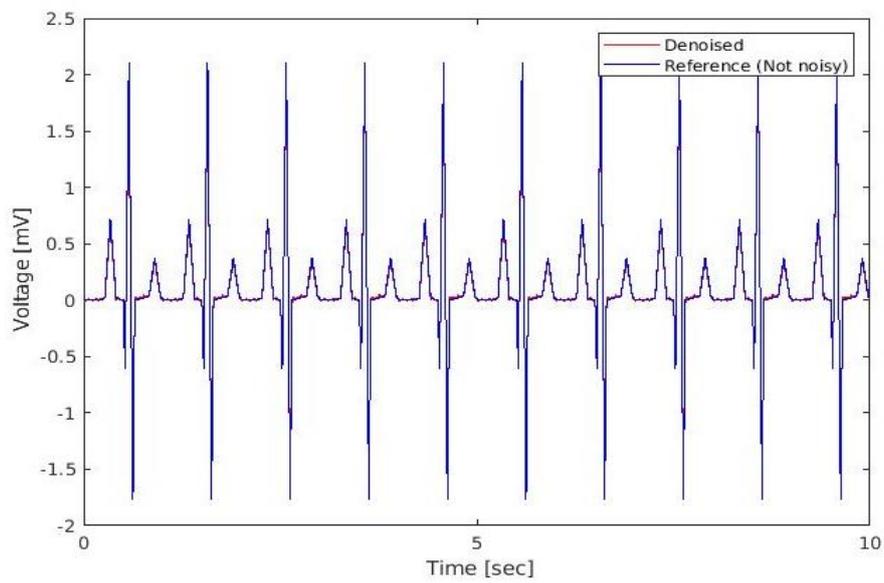

(b)

Figure 7. CNN model: a) ECG signal affected by strong drift and random noise. b) recovery of the original ECG signal (RMS = 0.0198).

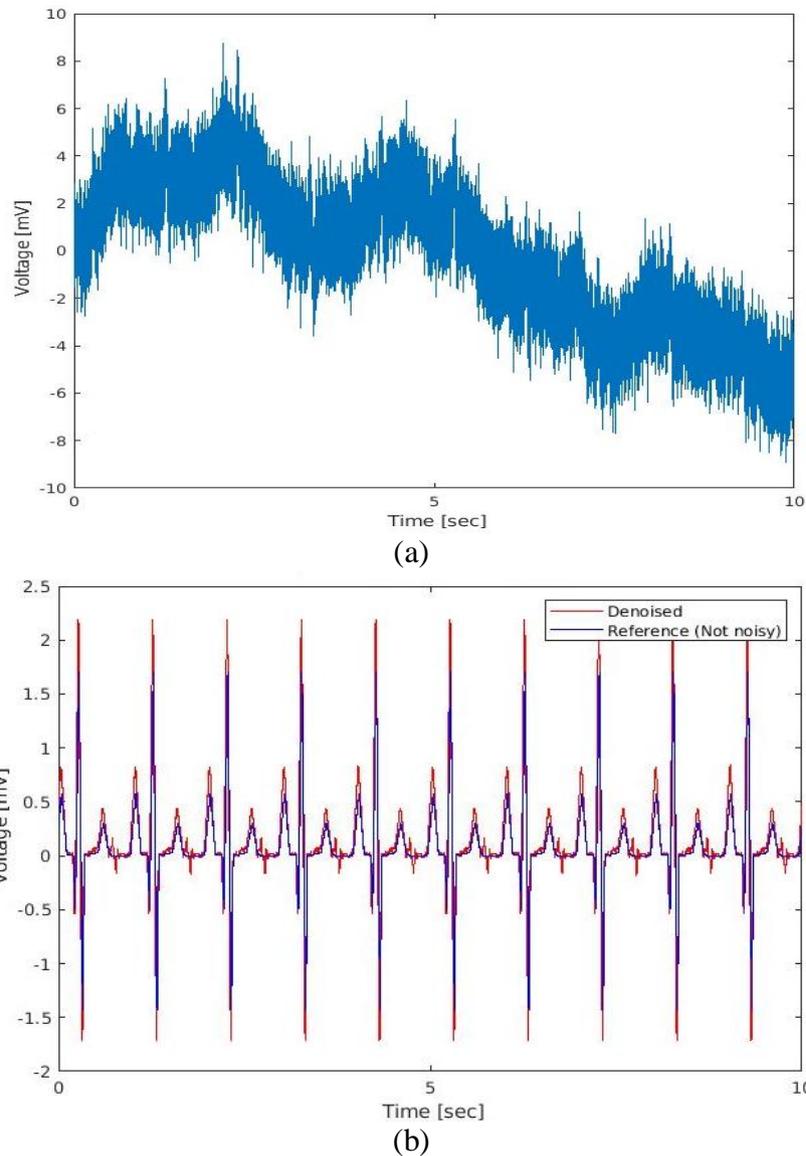

Figure 8. LSTM model: a) ECG signal affected by strong drift and random noise. b) recovery of the original ECG signal (RMS=0.2201).

The wavelet method was able to recover the original ECG signal in situations where the type of noise is random and additive and there is no drifting (RMS=0.1560), the wavelet RMS being about eight times higher than the RMS (0.0198) obtained with the CNN for the same ECG signal from Fig. 9.

The final result in Fig. 10 is obtained for the third dataset with real ECG signals with duration of 10 seconds. The CNN model is able to recover the original ECG signal (RMS=0.0220) while the noisy ECG signal contains synthetic generated noise and of the level produced by the electrode motion artefact. The average RMS over the testing dataset was 0.0625. The CNN model used was slightly different from the one described in Table 4 with only 5 convolutional layers and 36 filters per each convolutional layer and kernel size 9x1.

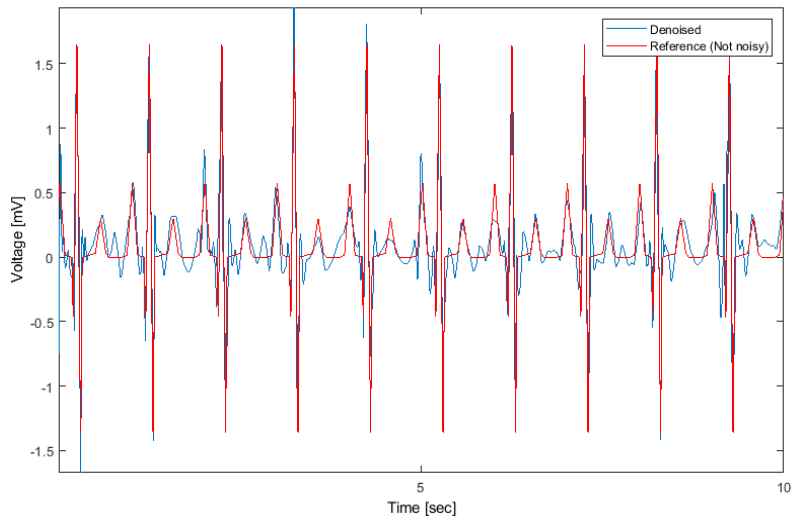

Figure 9. Wavelet model: recovery of the original ECG signal (RMS=0.1560).

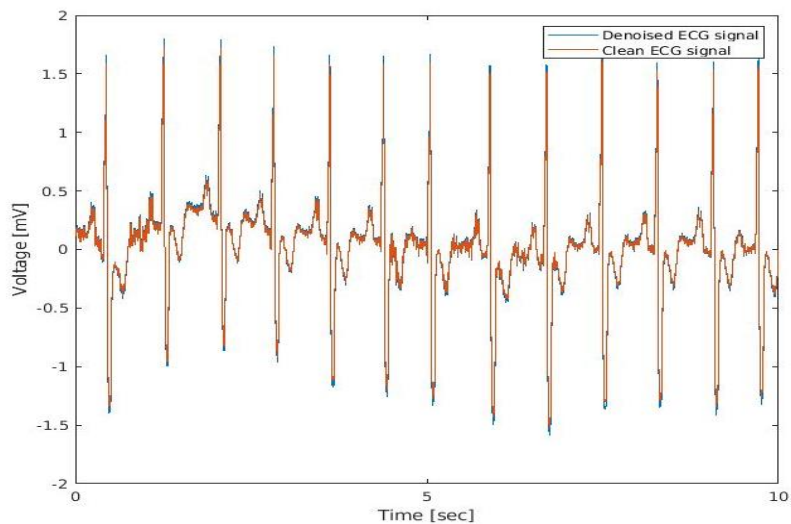

Figure 10. CNN model: recovery of the original ECG signal (RMS = 0.0220) from noisy ECG data affected by electrode motion artefact.

The model was left to converge for a long period of time (i.e. 331 minutes) so that to investigate the loss function which reached the value 14.1 after 1000 epochs (i.e. 14000 iterations) and which took about 5 hours (i.e. 331 minutes). Mini-batch size used for each training iteration was 400 where a mini-batch is a subset of the training set used to calculate the gradient of the loss function and update the weights. The loss function calculated over the mini-batch dataset was 13.6. The base learning rate used was 0.01 and the gradient threshold was 1.

The loss function is given by the optimization algorithm used by the DL model, which in this case is Adam. We mentioned already several times the Adam (Adaptive Moment Estimation) algorithm, which is used in order to update the weights of the CNN model and it is using the squared gradients to scale the learning rate and also the moving average of the gradient. In the mathematical literature there are many other optimization algorithms, which have been used in many and various engineering or bio-engineering optimization problems such as the Least Squares criterion [34, 35] or the Log-likelihood function [36].

The RMS calculated by the Adam algorithm for the mini-batch dataset is 5.22 for the same example shown in Fig. 10.

In conclusion, in this section the CNN models were proposed as regression models capable of rejecting very high levels of noise in the ECG signals, a situation, which has not been addressed before in the literature.

The results showed that the CNN model is superior to the LSTM model in the present settings (i.e. using DL toolbox of Matlab) both in the quality of results but also in the computational times: the CNN model took 53 minutes and 200 epochs to achieve better results for a testing dataset (i.e. average RMS =0.0279) than the LSTM model, which took 196 minutes and 2000 epochs (i.e. average RMS = 0.2321). In addition, the LSTM model required the generation of a second synthetic dataset containing four times more training input features with strong drift plus random noise in comparison to the CNN model training requirements. The quality of results was assessed visually and through the calculation of RMS errors.

It could be possible that by using other DL toolboxes (i.e. TensorFlow, Keras) available in other programming languages such as Python, to obtain slightly different results, as different DL toolboxes may offer further tuning of the various inner parameters of the CNN or LSTM models.

The results, which will be presented in the next sections, are first with regard to the promising performances of the DL models, which were obtained until now mainly on synthetic datasets with a limited interval of 57 to 67 heartbeats per minute. Therefore it is of interest to test the DL models with a larger heartbeat interval such as of 60 to 100. Second, it is paramount to investigate CNNs with shorter sequence input lengths so that to get closer to the real life situations where it is also needed to denoise ECG signals almost in real time, which implies the use of ECG sequences with one second length (i.e. a second by second study). Third, it will be also of interest to test further on real ECG datasets. Four, it is of interest also to compare the above DL models with other important DL models from literature such as Generative Adversarial Networks (GANs) or RBM.

## 4. *Design of Experiment study based on record 118 of MIT PhysioNet database*

All the real ECG datasets that will be used by the CNN model in the following sections, including in the Design of Experiment (DoE) study, will be filtered with the following algorithm [37]:

---

Algorithm 1. Initial filtering of the ECG signal:

1. Raw ECG data is low pass filtered with a frequency $F_l$=30 Hz.
2. The filtered ECG data is high pass filtering with a frequency $F_h$=0.1 Hz.
3. The filtered ECG data is notch filtered $F_d$= 47.5 Hz and $F_u$= 52.5 Hz using first order zero phase delay Butterworth filters (i.e. filtfilt in Matlab).
4. The baseline wander in the filtered ECG data is removed by using Wavelet Transform.

---

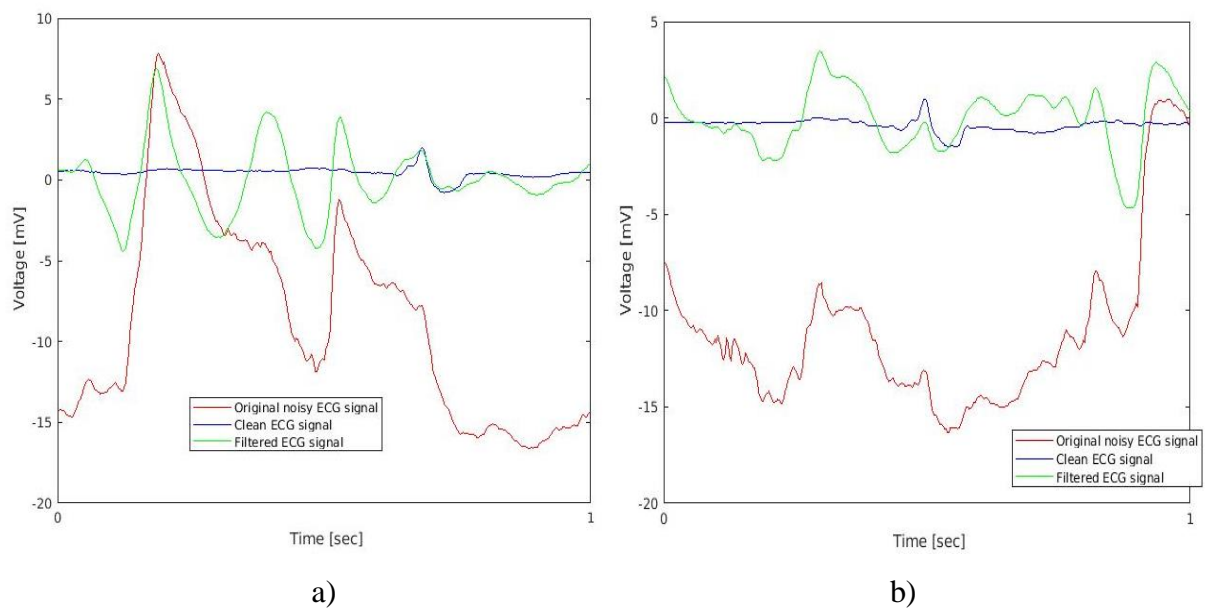

a)          b)

Figure 11. Comparison between the original noisy ECG signal from record 118 of MIT-BIH Arrhythmia database, the clean ECG signal and the ECG signal filtered with algorithm (1) from above: a) first example; b) second example.

In Fig. 11 is shown a comparison between the original noisy ECG signal from record 118 of MIT-BIH Arrhythmia database and the ECG signal filtered with algorithm (1) from above. It is clear that it is obtained an improvement in the quality of the ECG signal after the filtering implemented by algorithm (1), as the filtered ECG signal is closer to the clean ECG signal than the original noisy ECG signal, as it can be seen clearly visually.

Before going any further in our denoising task of ECG signals, we are first running a DoE study in which a CNN model with 3 convolutional layers is used with different number of filters in all the convolutional layers and with different kernel sizes for all the filters in all the convolutional layers of a CNN model.

The number of filters used in DoE were 16, 36, 56, 76, 96. The sizes of the kernels used were 9x1, 13x1, 23x1, 33x1, 43x1, 53x 1, 83x1, 99x1, 9x9, 23x23, 45x45. The mini-batch size for the data was 200, initial learning rate 0.01 and gradient threshold 1 and with Adam optimizer. The ECG signals in this case have 1 second in length with 360 sample points. The general structure of the CNN model is formed of an Image Input, Convolution Layer, Batch Normalization, ReLu, Average Pooling, Convolutional Layer, Batch Normalization, ReLu, Average Pooling, Convolutional Layer, Batch Normalization, ReLu, Average Pooling, Fully Connected, Regression output. The average pooling layer of size 1 and stride [4 1] (i.e. sub-sampling).

The scope of running a DoE is to investigate what is the best structure of the CNN model for denoising ECG signals that is the number of filters, the kernel sizes or the number of convolutional layers (i.e. parameter not included in this DoE study). Further work will be to test with more filters, kernel sizes and to include also the number of layers in the DoE.

The dataset used has 2880 ECG signals from record 118 (i.e. SNRs available SNR=-6, SNR=0, SNR=6, SNR=12) with 720 different ECG patterns for each SNR available. Each pattern has a duration of 1 second. The entire dataset is divided in four parts, from which three parts represent the training dataset (i.e. randomly shuffled) and one part is the testing dataset (i.e. randomly shuffled).

In total there were run 51 simulations and each simulation consisted of 2000 epochs. In Table 3 is summarized the average RMS and the average SNR values calculated over the testing datasets together with the computational times for running the 2000 epochs. From looking quickly in the table below it can be observed that with more filters in a layer and/or bigger kernel sizes the accuracy of results increased (i.e. smaller RMS, higher SNR) but also the computational times increased. The table is more simple summarized in Fig. 12.

Table 3. DoE study for determining the best structure of CNN model (i.e. 3 convolutional layers) with regard to accuracy of results and computational time after 2000 epochs.

| Nr. | Number of filters per conv layer | Kernel size | Average RMS value [mV] | Average SNR value [dB] | Computational time (2000 epochs) [seconds] |
|---|---|---|---|---|---|
| 1 | 16 | 9x1 | 0.2367 | 6.0378 | 559 |
| 2 | 16 | 13x1 | 0.2158 | 6.9879 | 560 |
| 3 | 16 | 23x1 | 0.1972 | 7.8218 | 602 |
| 4 | 16 | 33x1 | 0.1818 | 8.4814 | 657 |
| 5 | 16 | 43x1 | 0.1762 | 8.7292 | 1445 |
| 6 | 16 | 53x1 | 0.1691 | 9.0137 | 1526 |
| 7 | 16 | 83x1 | 0.1707 | 8.9353 | 1965 |
| 8 | 16 | 99x1 | 0.1763 | 8.6390 | 2259 |
| 9 | 16 | 9x9 | 0.2302 | 6.3339 | 561 |
| 10 | 16 | 23x23 | 0.2013 | 7.6448 | 645 |
| 11 | 16 | 45x45 | 0.1616 | 9.1224 | 1478 |
| 12 | 36 | 9x1 | 0.1655 | 9.5205 | 547 |
| 13 | 36 | 13x1 | 0.1585 | 9.8539 | 667 |
| 14 | 36 | 23x1 | 0.1343 | 11.3061 | 1313 |
| 15 | 36 | 33x1 | 0.1342 | 11.3076 | 2161 |
| 16 | 36 | 43x1 | 0.1366 | 11.1338 | 9620 |
| 17 | 36 | 53x1 | 0.1268 | 11.7833 | 11137 |
| 18 | 36 | 83x1 | 0.1406 | 10.7736 | 14870 |
| 19 | 36 | 99x1 | 0.1352 | 11.1137 | 22256 |
| 20 | 36 | 9x9 | 0.1678 | 9.3044 | 2442 |
| 21 | 36 | 23x23 | 0.1414 | 10.7839 | 3259 |
| 22 | 36 | 45x45 | 0.1429 | 10.6882 | 14423 |
| 23 | 56 | 9x1 | 0.1411 | 10.9956 | 1706 |
| 24 | 56 | 13x1 | 0.1339 | 11.3814 | 1978 |
| 25 | 56 | 23x1 | 0.1395 | 11.0942 | 3147 |
| 26 | 56 | 33x1 | 0.1338 | 11.5501 | 5184 |
| 27 | 56 | 43x1 | 0.1303 | 11.7866 | 11876 |
| 28 | 56 | 53x1 | 0.1360 | 11.2322 | 13107 |
| 29 | 56 | 83x1 | 0.1408 | 10.9639 | 32417 |
| 30 | 56 | 99x1 | 0.1324 | 11.5617 | 16256 |
| 31 | 56 | 9x9 | 0.1404 | 11.0083 | 618 |
| 32 | 56 | 23x23 | 0.1259 | 11.7115 | 1137 |
| 33 | 56 | 45x45 | 0.1287 | 11.6914 | 15506 |
| 34 | 76 | 9x1 | 0.1318 | 11.6463 | 755 |
| 35 | 76 | 13x1 | 0.1304 | 11.5585 | 958 |
| 36 | 76 | 23x1 | 0.1273 | 11.9067 | 1812 |
| 37 | 76 | 33x1 | 0.1302 | 11.7406 | 6490 |
| 38 | 76 | 43x1 | 0.1237 | 12.3038 | 11578 |
| 39 | 76 | 53x1 | 0.1303 | 11.6756 | 16137 |
| 40 | 76 | 83x1 | 0.1330 | 11.3440 | 30671 |
| 41 | 76 | 99x1 | 0.1285 | 11.9661 | 38940 |
| 42 | 76 | 9x9 | 0.1365 | 11.2288 | 764 |
| 43 | 76 | 23x23 | 0.1285 | 11.8275 | 1810 |
| 44 | 76 | 45x45 | 0.1351 | 11.3557 | 10371 |
| 45 | 96 | 9x1 | 0.1368 | 11.3084 | 850 |
| 46 | 96 | 13x1 | 0.1235 | 12.3622 | 1142 |
| 47 | 96 | 23x1 | 0.1235 | 12.4544 | 2234 |
| 48 | 96 | 33x1 | 0.1239 | 12.2720 | 5911 |
| 49 | 96 | 43x1 | 0.1284 | 11.9655 | 15638 |
| 50 | 96 | 53x1 | 0.1310 | 11.8863 | 19206 |
| 51 | 96 | 83x1 | 0.1384 | 11.0808 | 39150 |

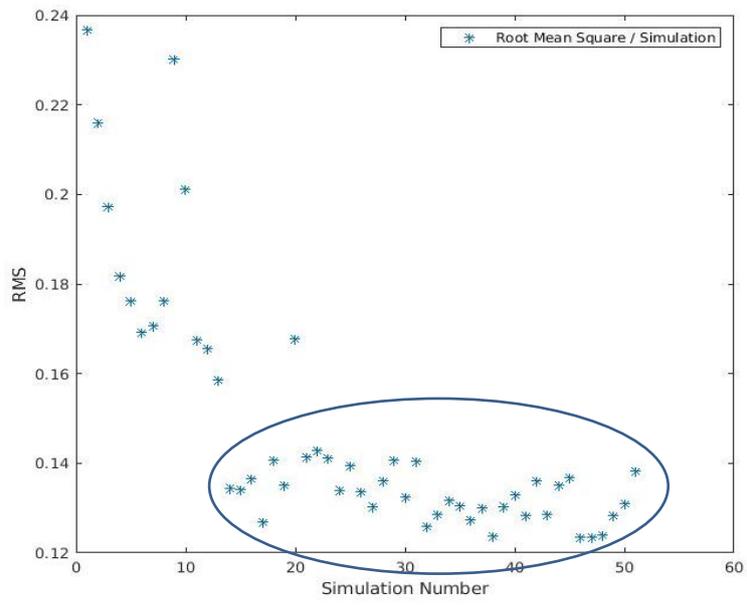

a)

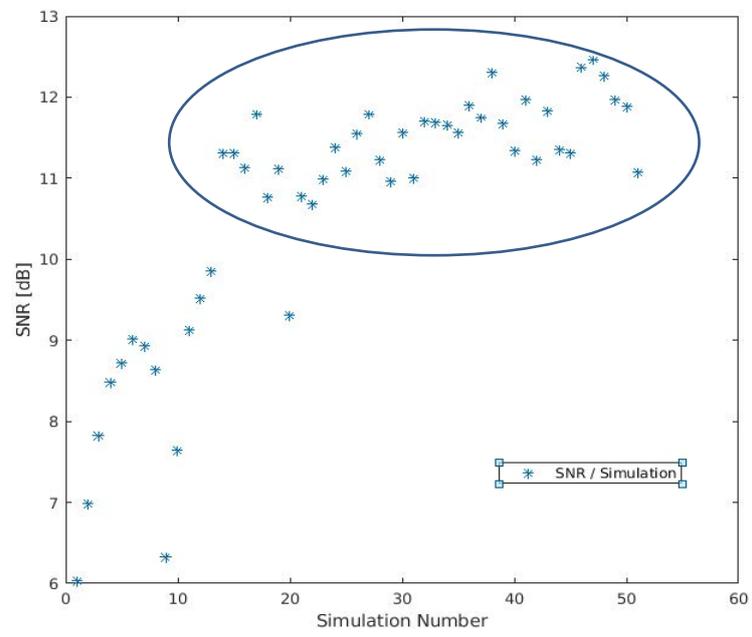

b)

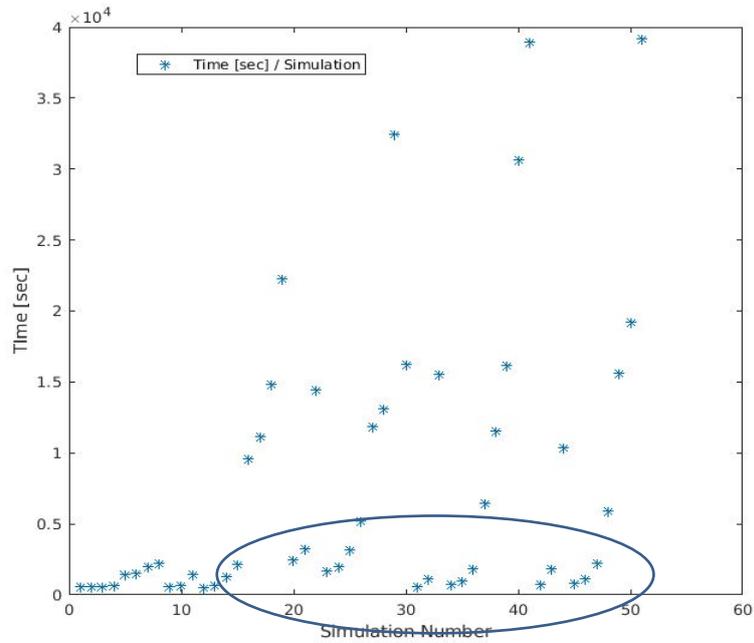

c)

Figure 12. Average RMS, average SNR, computational times for 51 simulations and different CNN architectures: a) average RMS calculated over the testing dataset; b) average SNR calculated over the testing dataset; c) computational times.

By looking to Fig 12(a) and Fig 12(b), it can be noticed that the simulations beyond simulation 13 are of interest because RMS values are low and SNR values tend to be high. Now by looking to Fig 12(c), it can be noticed that from all the simulations beyond simulation 13 and up to simulation 51, the simulations with computational times smaller than approximately 5000 seconds (i.e. threshold) are of interest. This results in a subset of 18 simulations which are 14, 15, 20, 21, 23, 24, 25, 26, 31, 32, 34, 35, 36, 42, 43, 45, 46 and 47. The rest of simulations have computational times higher than approximately 5000 seconds while the RMS and SNR values are either worse or quite similar to this subset of 18 simulations.

Table 4. The 18 simulations with the best RMS and SNR values and the shortest computational times.

| Nr. | Number of filters per conv layer | Kernel size | Average RMS value [mV] | Average SNR value [dB] | Computational time (2000 epochs) [seconds] |
|---|---|---|---|---|---|
| 14 | 36 | 23x1 | 0.1343 | 11.3061 | 1313 |
| 15 | 36 | 33x1 | 0.1342 | 11.3076 | 2161 |

| 20 | 36 | 9x9 | 0.1678 | 9.3044 | 2442 |
| **21** | **36** | **23x23** | **0.1414** | **10.7839** | **3259** |
| 23 | 56 | 9x1 | 0.1411 | 10.9956 | 1706 |
| 24 | 56 | 13x1 | 0.1339 | 11.3814 | 1978 |
| 25 | 56 | 23x1 | 0.1395 | 11.0942 | 3147 |
| <span style="color:red">**26**</span> | <span style="color:red">**56**</span> | <span style="color:red">**33x1**</span> | <span style="color:red">**0.1338**</span> | <span style="color:red">**11.5501**</span> | <span style="color:red">**5184**</span> |
| 31 | 56 | 9x9 | 0.1404 | 11.0083 | 618 |
| 32 | 56 | 23x23 | 0.1259 | 11.7115 | 1137 |
| 34 | 76 | 9x1 | 0.1318 | 11.6463 | 755 |
| 35 | 76 | 13x1 | 0.1304 | 11.5585 | 958 |
| 36 | 76 | 23x1 | 0.1273 | 11.9067 | 1812 |
| 42 | 76 | 9x9 | 0.1365 | 11.2288 | 764 |
| 43 | 76 | 23x23 | 0.1285 | 11.8275 | 1810 |
| 45 | 96 | 9x1 | 0.1368 | 11.3084 | 850 |
| <span style="color:green">**46**</span> | <span style="color:green">**96**</span> | <span style="color:green">**13x1**</span> | <span style="color:green">**0.1235**</span> | <span style="color:green">**12.3622**</span> | <span style="color:green">**1142**</span> |
| 47 | 96 | 23x1 | 0.1235 | 12.4544 | 2234 |

Based on Table 4 and the 18 simulations with the best RMS and SNR values and the shortest computational times, it results that the optimal CNN model would be simulation 46 (i.e. green colour) that is the simulation based on the CNN model with 3 convolutional layers, 96 filters for each layer and kernel size 13x1: in this simulation the computational time was only 1142 seconds while the average RMS over the testing dataset was 0.1235 and the average SNR [dB] over the testing dataset was 12.3622.

Similarly, based on the same Table 4, the worst simulation from the point of view of the computational time, corresponded to simulation 26 (i.e. red colour) for a CNN model with 56 filters in a convolutional layer, 3 convolutional layers and kernel size 33x1: the RMS was 0.1338, the SNR was 11.5501 dB and the computational time was the biggest from all the simulations that is 5184 seconds.

In the remaining part of this work, it will be used an under-optimal structure of a CNN model especially from the computational time point of view and which corresponds to simulation 21 (i.e. black bolded font) for a CNN architecture with 3 convolutional layers, 36 filters per convolutional layer and kernel size 23x23. This architecture produced an average RMS over the testing dataset of 0.1414 and an average SNR [dB] over the testing dataset of 10.7839. These values are quite close to the ones obtained with the optimal CNN model but the computational time was bigger than the one obtained with the optimal structure that is 3259 seconds as compared to only 1142 seconds. Future work it will involve using the optimal CNN architecture as devised in Table 4 (i.e. simulation 46 - green colour) and comparing it with this under-optimal CNN model.

The type of CNN model used in the remaining section of this paper is a CNN model with 3 convolutional layers, 36 filters per layer and an average pooling layer of size 1 and stride [4 1] (i.e. sub-sampling) (Table 5). The mini-batch size for the data was 200, initial learning rate 0.01 and gradient threshold 1 and with Adam optimizer. The ECG signals in this case have 1 second in length with 360 sample points.

TABLE 5. LISTING OF THE CNN LAYERS: $M = 360$ IS THE NUMBER OF SAMPLES PER INPUT ECG SIGNAL WITH DURATION OF 1 SECOND

| Nr | Type | Description |
|---|---|---|
| 1 | Image Input | 360x1x1 images with 'zerocenter' normalization |
| 2 | Convolution | 36 23x23x1 convolutions with stride [1 1] and padding 'same' |
| 3 | Batch Normalization | Batch normalization with 36 channels |
| 4 | ReLu | Rectified Linear Unit |
| 5 | Average Pooling | 1x1 average pooling with stride [4 1] and padding [0 0 0 0] |
| 6 | Convolution | 36 23x23x1 convolutions with stride [1 1] and padding 'same' |
| 7 | Batch Normalization | Batch normalization with 36 channels |
| 8 | ReLu | Rectified Linear Unit |
| 9 | Average Pooling | 1x1 average pooling with stride [4 1] and padding [0 0 0 0] |
| 10 | Convolution | 36 23x23x1 convolutions with stride [1 1] and padding 'same' |
| 11 | Batch Normalization | Batch normalization with 36 channels |
| 12 | ReLu | Rectified Linear Unit |
| 13 | Average Pooling | 1x1 average pooling with stride [4 1] and padding [0 0 0 0] |
| 14 | Fully Connected | 360 fully connected layer |
| 15 | Regression Output | Mean-squared-error with response 'Response' |

## 4.1  Train and test on record 118 of MIT PhysioNet database

This section presents results from training and testing on record 118 of MIT PhysioNet database and from a single channel/electrode. The same as above, the dataset has 2880 ECG signals (i.e. SNRs available SNR=-6, SNR=0, SNR=6, SNR=12) with 720 different ECG patterns for each SNR available. Each pattern has a duration of 1 second. The entire dataset is divided in four parts, from which three parts represent the training dataset and one part is the testing dataset.

The training of the CNN model took 1 hour and 32 minutes for 6000 epochs (i.e. 60000 iterations) to reach RMS for the mini-batch data of 1.14 and also mini-batch loss of 0.6. Over the entire testing dataset of size 720 input ECG distinct features, the average RMS over the entire testing dataset is 0.0871 and the average SNR over the entire testing dataset is 15.02.

In Fig. 13 there are shown denoised ECG signals (i.e. duration 1 second) with CNN model for different levels of SNR between the initial noisy ECG and clean ECG signals: in Fig. 13(a) the initial SNR between the reference (not noisy) ECG signal and the noisy ECG signal is -6, while the RMS between the denoised ECG signal and the clean ECG signal is 0.12 and the SNR between the denoised ECG signal and the clean ECG signal is 9.76.

In Fig. 13(b) the initial SNR between the reference (not noisy) ECG signal and the noisy ECG signal is -6, while the RMS between the denoised ECG signal and the clean ECG signal is 0.16 and the SNR between the denoised ECG signal and the clean ECG signal is 8.96.

In Fig. 13(c) the initial SNR between the reference (not noisy) ECG signal and the noisy ECG signal is 0, while the RMS between the denoised ECG signal and the clean ECG signal is 0.10 and the SNR between the denoised ECG signal and the clean ECG signal is 11.74.

Finally, in Fig. 13(d) the initial SNR between the reference (not noisy) ECG signal and the noisy ECG signal is 0, while the RMS between the denoised ECG signal and the clean ECG signal is 0.09 and the SNR between the denoised ECG signal and the clean ECG signal is 12.78.

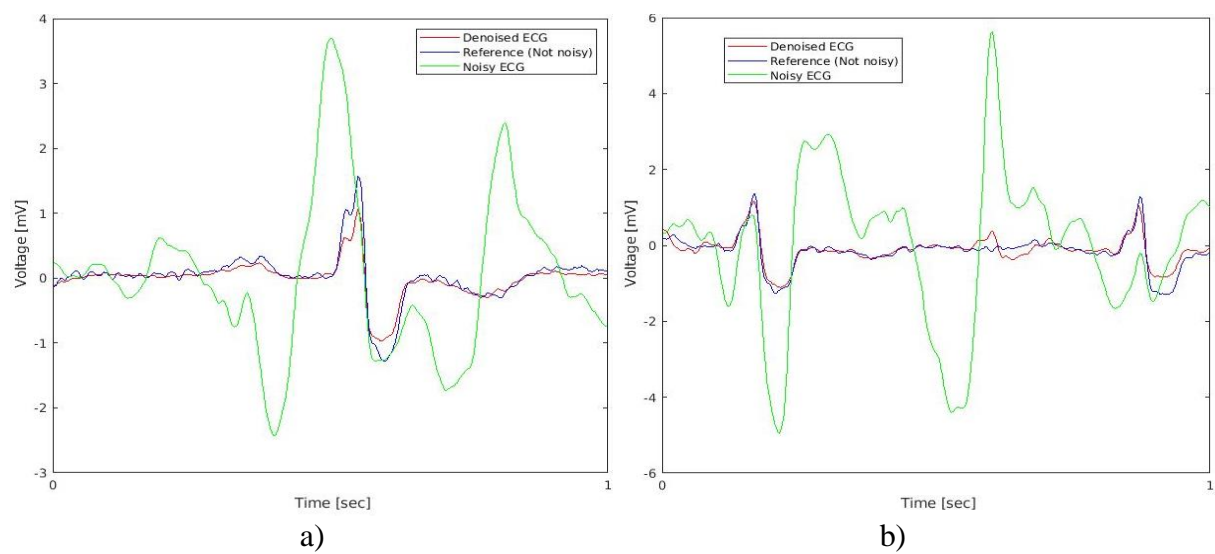

a)                                    b)

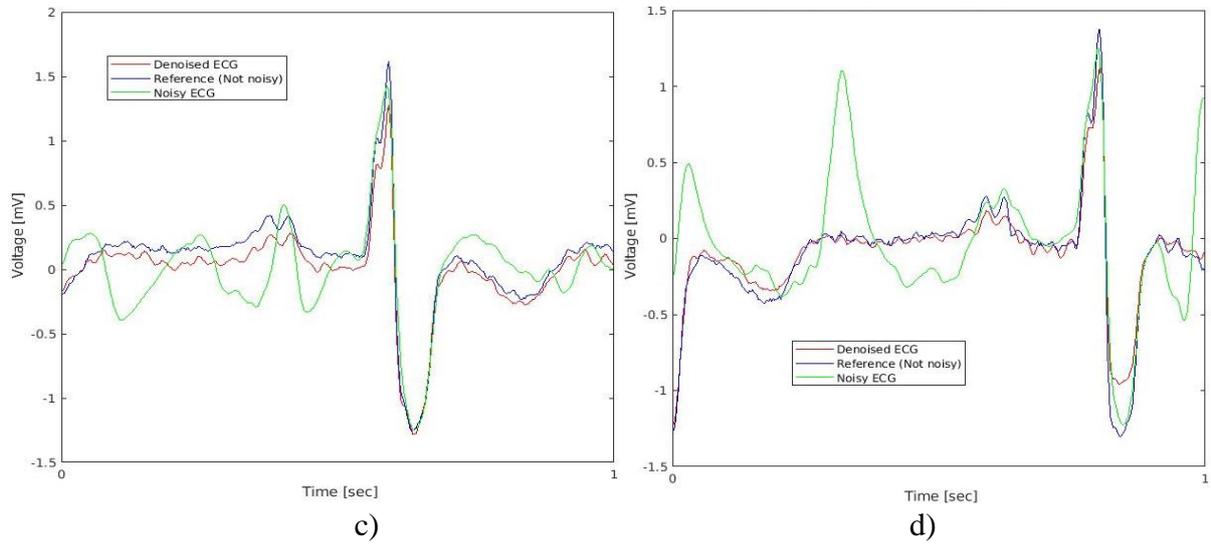

Figure 13. Denoised ECG signals with CNN model for different levels of SNR between the initial noisy ECG and clean ECG signals: a) RMS = 0.12, SNR = 9.76; b) RMS = 0.16, SNR = 8.96 ; c) RMS = 0.10, SNR = 11.74 ; d) RMS = 0.09, SNR = 12.78.

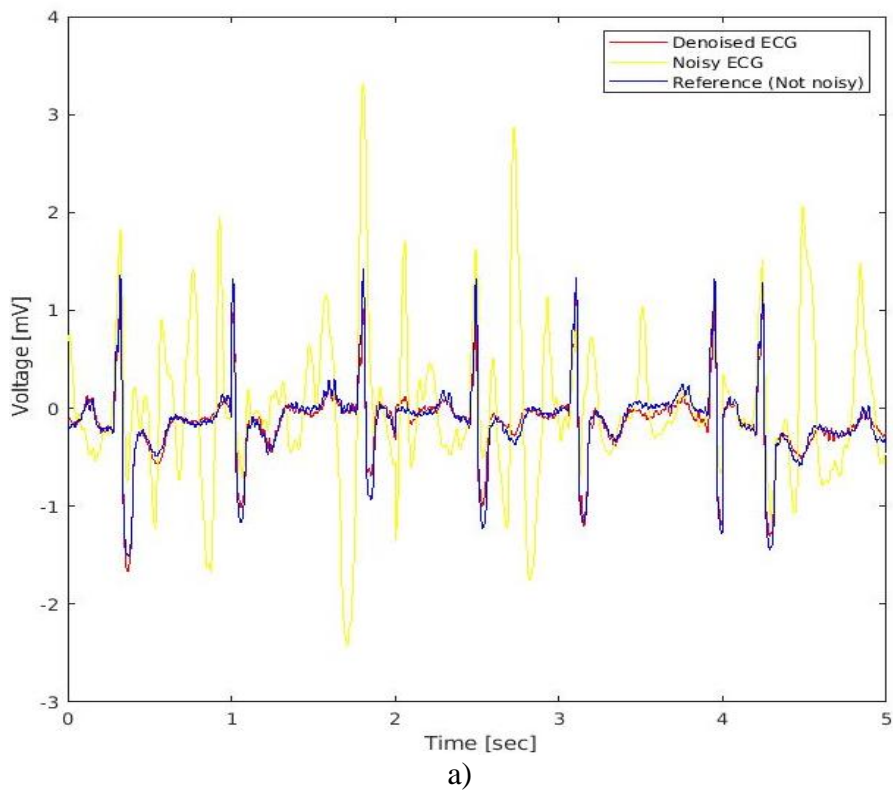

a)

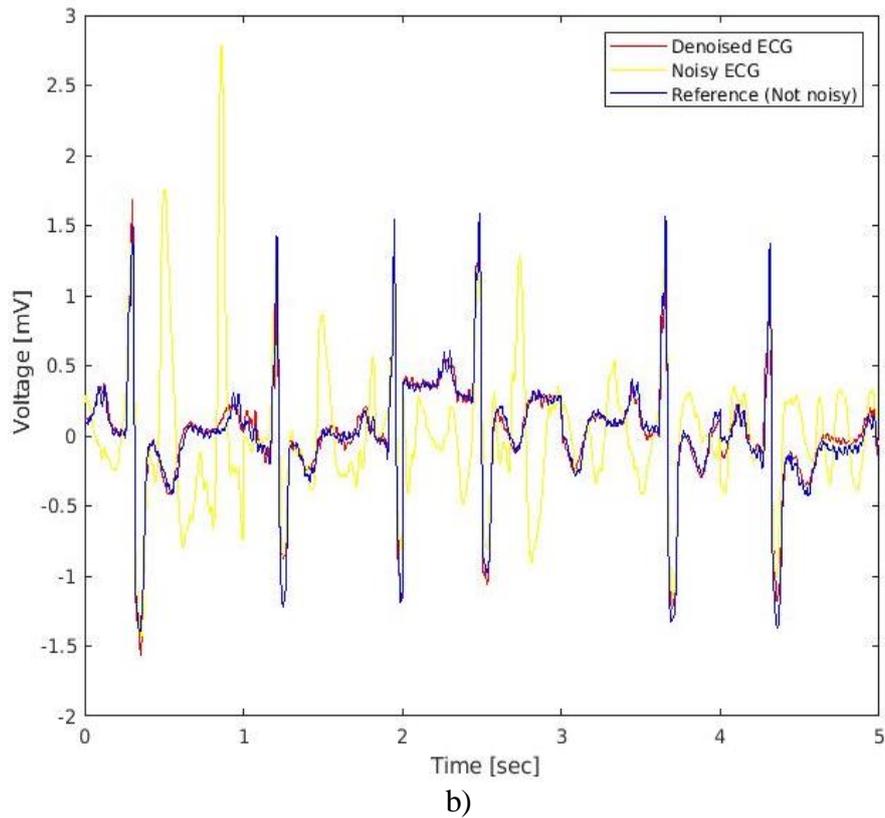

b)

Figure 14. Sequence of 5 seconds denoised ECG signals with CNN model for different levels of SNR between the initial noisy ECG and clean ECG signals: a) RMS = 0.0840, SNR = 14.03; b) RMS = 0.0712, SNR = 15.68.

Finally, in Fig. 14(a) is shown a sequence of 5 seconds denoised ECG signal. The noisy ECG signal shown in the figure was initially filtered with the conventional methods discussed above (i.e. algorithm 1). The denoised ECG signal obtained with the CNN model has an average RMS over the 5 seconds of 0.0840 and the average SNR over the 5 seconds of 14.03 between the denoised and the clean ECG signal.

In Fig. 14(b) is shown a sequence of 5 seconds denoised ECG signal. The noisy ECG signal shown in the figure was initially filtered with the conventional methods discussed above (i.e. algorithm 1). The denoised ECG signal obtained with the CNN model has an average RMS over the 5 seconds of 0.0712 and the average SNR over the 5 seconds of 15.68 between the denoised and the clean ECG signal.

## 4.2 Train and test on record 105 of MIT PhysioNet database

This section presents results from training and testing on record 105 of MIT PhysioNet database. The model used is a CNN model with 3 convolutional layers, 36 filters and an average pooling layer of size 1 and stride [4 1] (i.e. sub-sampling) (Table 6). The mini-batch size for the data was 200, initial learning rate 0.01 and gradient threshold 1 and with Adam optimizer. The ECG signals in this case have 1 second in length with 360 sample points.

The dataset has in total 7920 ECG signals (i.e. SNRs available SNR =-8, SNR=-6, SNR=-3, SNR=-1, SNR=0, SNR=3, SNR=6, SNR=8, SNR=12, SNR=20, SNR=36) with ECG signal duration of 1 second and which dataset is divided in four parts from which three parts represent the training dataset and one part is the testing dataset.

TABLE 6. LISTING OF THE CNN LAYERS: $M = 360$ IS THE NUMBER OF SAMPLES PER INPUT ECG SIGNAL WITH DURATION OF 1 SECOND

| Nr | Type | Description |
| --- | --- | --- |
| 1 | Image Input | 360x1x1 images with 'zerocenter' normalization |
| 2 | Convolution | 36 23x23x1 convolutions with stride [1 1] and padding 'same' |
| 3 | Batch Normalization | Batch normalization with 36 channels |
| 4 | ReLu | Rectified Linear Unit |
| 5 | Average Pooling | 1x1 average pooling with stride [4 1] and padding [0 0 0 0] |
| 6 | Convolution | 36 23x23x1 convolutions with stride [1 1] and padding 'same' |
| 7 | Batch Normalization | Batch normalization with 36 channels |
| 8 | ReLu | Rectified Linear Unit |
| 9 | Average Pooling | 1x1 average pooling with stride [4 1] and padding [0 0 0 0] |
| 10 | Convolution | 36 23x23x1 convolutions with stride [1 1] and padding 'same' |
| 11 | Batch Normalization | Batch normalization with 36 channels |
| 12 | ReLu | Rectified Linear Unit |
| 13 | Average Pooling | 1x1 average pooling with stride [4 1] and padding [0 0 0 0] |
| 14 | Fully Connected | 360 fully connected layer |
| 15 | Regression Output | Mean-squared-error with response 'Response' |

Since it is of interest to use such applications in real-life or near real-life situations, the training of the CNN model was done for only 400 epochs which took 18 minutes (i.e. 11600 iterations) to reach RMS for the mini-batch data of 1.25 and also mini-batch loss of 0.8.

Over the entire testing dataset of size 1980 the average RMS over the entire testing dataset is 0.0869 and the average SNR [dB] over the entire testing dataset is 14.77.

Furthermore, the number of filters in each layer is increased to 72 and the above simulation is repeated. The training of the CNN model was done again for only 400 epochs which took 20 minutes (i.e. 11600 iterations) to reach RMS for the mini-batch data of 1.08 (i.e. similar as before) and also mini-batch loss of 0.6 (i.e. similar as before).

However, over the entire testing dataset of size 1980 the average RMS over the entire testing dataset is 0.0729 and the average SNR over the entire testing dataset is 16.45 and these numerical results are slightly better than in the first simulation, which means that increasing the number of filters in a layer may have the positive effect of increasing the accuracy of results but the negative side is the increasing of the computational time (i.e. 2 minutes more). This finding is more generally confirmed also by the previous DoE study.

In Fig. 15 there are shown denoised ECG signals (i.e. duration 1 second) with CNN model for different levels of SNR between the initial noisy ECG and clean ECG signals: in Fig. 15 (a) the initial SNR between the reference (not noisy) ECG signal and the noisy ECG signal is -6, while the RMS between the denoised ECG signal and the clean ECG signal is 0.055 and the SNR between the denoised ECG signal and the clean ECG signal is 14.99.

In Fig. 15(b) the initial SNR between the reference (not noisy) ECG signal and the noisy ECG signal is -6, while the RMS between the denoised ECG signal and the clean ECG signal is 0.066 and the SNR between the denoised ECG signal and the clean ECG signal is 13.30.

In Fig. 15(c) the initial SNR between the reference (not noisy) ECG signal and the noisy ECG signal is 0, while the RMS between the denoised ECG signal and the clean ECG signal is 0.098 and the SNR between the denoised ECG signal and the clean ECG signal is 21.04. Finally, in Fig. 15(d) the initial SNR between the reference (not noisy) ECG signal and the noisy ECG signal is 0, while the RMS between the denoised ECG signal and the clean ECG signal is 0.05 and the SNR between the denoised ECG signal and the clean ECG signal is 15.54.

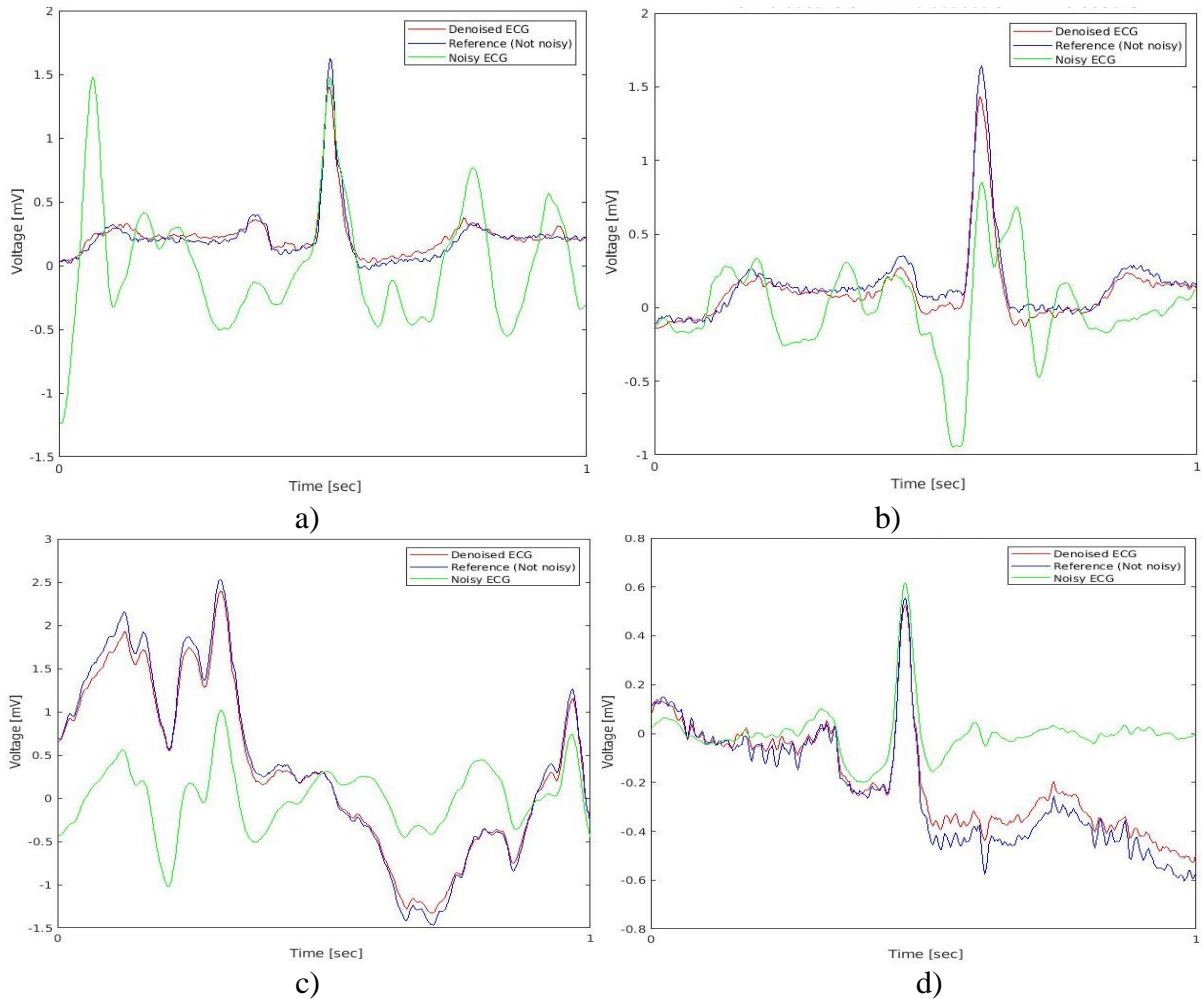

Figure 15. Denoised ECG signals with CNN model for different levels of SNR between the initial noisy ECG and clean ECG signals: a) RMS = 0.055, SNR = 14.99; b) RMS = 0.066, SNR = 13.30 ; c) RMS = 0.098 , SNR = 21.04; d) RMS = 0.05, SNR = 15.54.

In Fig.16(a) is shown a sequence of 5 seconds denoised ECG signal. The noisy ECG signal shown in the figure was initially filtered with the conventional methods discussed above. The denoised ECG signal obtained with the CNN model has an average RMS over the 5 seconds of 0.0454 and the average SNR over the 5 seconds of 18.60 between the denoised and clean ECG signal. In Fig. 16(b) is shown a sequence of 5 seconds denoised ECG signal. The noisy ECG signal shown in the figure was initially filtered with the conventional methods discussed above (i.e. algorithm 1). The denoised ECG signal obtained with the CNN model has an average RMS over the 5 seconds of 0.0416 and the average SNR [dB] over the 5 seconds of 18.32 between the denoised and clean ECG signal.

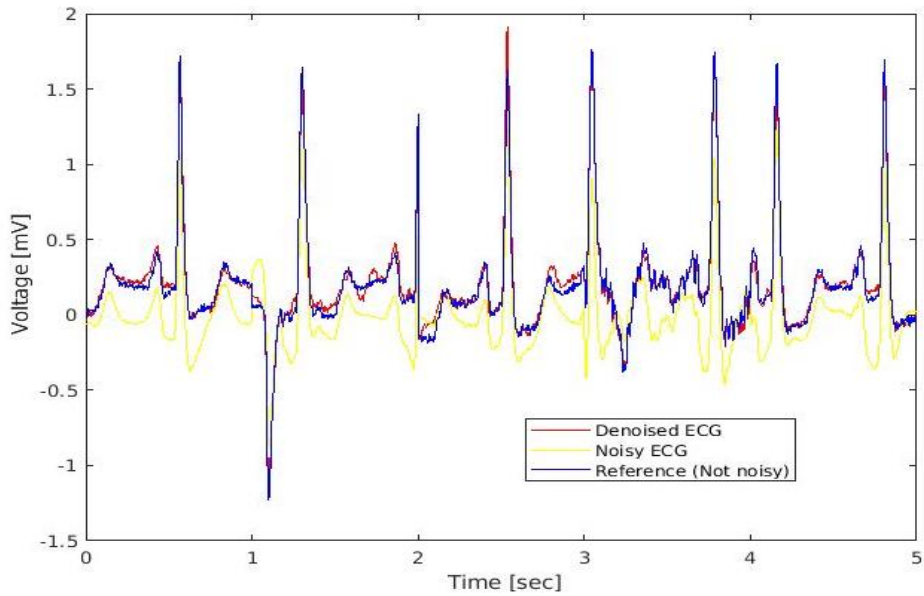

a)

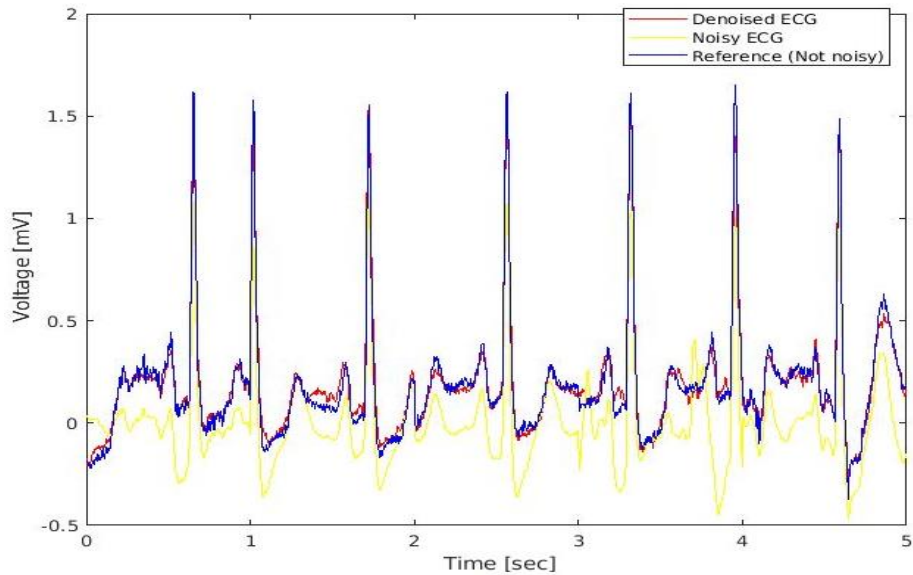

b)

Figure 16. Sequence of 5 seconds denoised ECG signals with CNN model for different levels of SNR between the initial noisy ECG and clean ECG signals: a) RMS = 0.454, SNR = 18.60; b) RMS = 0.0416, SNR = 18.32.

## *4.3 Comparison of CNN model with Restricted Boltzmann Machine model on MIT PhysioNet Database*

In this section it is performed a comparison between the CNN model and a RBM applied on two records of the MIT PhysioNet database.

The RBM consists of one input/visible layer ($i_1,\ldots, i_6$), one hidden layer ($h_1, h_2$) and the corresponding biases vectors that are bias a and bias b, as well as the weights connecting the two layers. The typical architecture is shown in Fig. 17.

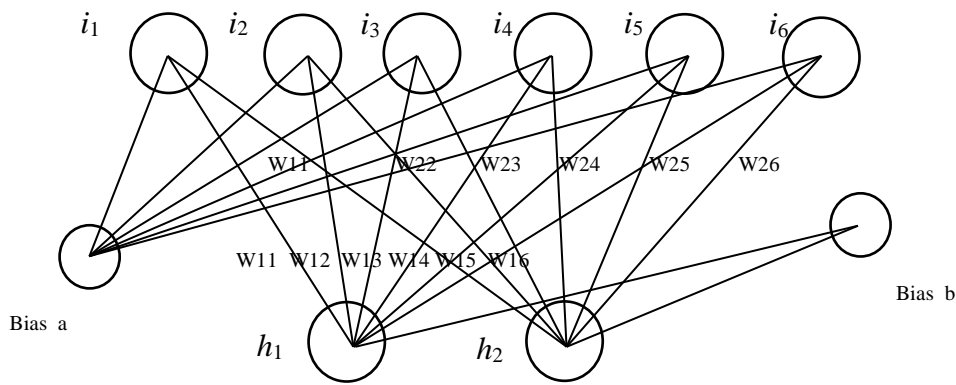

Figure 17. Typical architecture of RBM.

It is attempted to minimize the energy function, which is defined as:

$$E(i,h) = -\sum_k a_k i_k - \sum_m b_m h_m - \sum_{km} i_m h_m w_{km} \qquad (5)$$

It is observed that the value of the energy function depends on the configurations of visible/input states, hidden states, weights and biases. Furthermore, the nature of the RBM is probabilistic. The model assigns probabilities and at each moment the RBM is in a state. The probability that a specific state for *i* and *h* can be detected is calculated with the following joint probability distribution:

$$P(i,h) = \frac{1}{T} e^{-E(i,h)}$$
$$T = \sum_{i,h} e^{-E(i,h)} \qquad (6)$$

where *T* is a summation over all possible pairs of input/visible and hidden vectors.

It is possible to calculate the conditional probability for a hidden neuron *s* being activated (i.e. 1) as well as the probability that the input neuron *t* is set to 1:

$$P(h_s = 1|i) = \frac{1}{1+e^{(-(b_s+w_s i_t))}} = \sigma(b_s + \sum_t i_s w_{st})$$

$$P(i_t = 1|h) = \frac{1}{1+e^{(-(a_t+w_t h_s))}} = \sigma(a_t + \sum_s h_s w_{st})$$

(7)

where σ is the Sigmoid function.

Now the weights of the model are updated iteratively by using for example Gibbs sampling so that to reach the minimum for the energy function from eq.(5).

In this work, the structure of the RBM is fully described in [13] and it is based also on the work described in [38]: a stack of three RBM layers is used, each of the layer with about 1000 hidden neurons (i.e. 1024, 1024, 1280). The model is trained with sequences of ECG signals of length 2 seconds and data from two channels is used, which will make it harder for the model to train. Each RBM layer is trained for 100 epochs. Within each epoch the training data is divided in a number of 17975 batches (i.e. each with size 32) and for each batch is calculated an error. The RBM model is implemented in Octave programming language [13][38] and it was trained on an Intel i5. Therefore it took several days (i.e. 4 days) to accomplish.

For comparison purposes, a CNN model similar to the one described in Table 6 is trained with the ECG signals from record 105, with data from two channels (i.e. similar to the training of RBM) and each ECG signal with a length of a second.

In Table 7 it is shown a comparison between the RBM and the CNN model for the RMS values obtained for the same record 105 and for a testing set of signals and for different SNR values. The average RMS for the CNN model of 0.0833 seems to outperform the RBM with an average RMS of 0.2286. A similar comparison is shown in Table 8 for record 118, for a testing dataset and for RBM and CNN model and for different SNR values. This time the difference is for the average RMS over the testing dataset for the CNN model of 0.1192 and the average RMS for the RBM over a testing set of signals of 0.3258. In these settings, the CNN model seemed better than the RBM model although further investigations are necessary such as training/testing the RBM model also with 1 second ECG signals.

In Fig. 18, it can be seen a comparison between the denoised/reconstructed ECG signal, the original/clean ECG signal and the noisy ECG signal for SNR = -6 and for record 105 of MIT PhysioNet database with the following RMS values calculated between the denoised and the clean ECG signals: Fig. 18(a) RMS=0.1799, Fig. 18(b) RMS = 0.1650, Fig. 18(c) RMS =0.15506, Fig. 18(d) RMS =0.14035.

Table 7. Comparison of RMS values obtained for record 105 and for testing dataset for RBM model and CNN model and for different SNR values.

| SNR | RBM model (i.e. 2 seconds training) | CNN model (i.e. 1 second training) |
|---|---|---|
| -8 | 0.463 | 0.0613 |
| -6 | 0.192 | 0.1205 |
| 0 | 0.171 | 0.084 |
| 6 | 0.162 | 0.082 |
| 12 | 0.157 | 0.069 |

Table 8. Comparison of RMS values obtained for record 118 and for testing dataset for RBM model and CNN model and for different SNR values.

| SNR | RBM model (i.e. 2 seconds training) | CNN model (i.e. 1 second training) |
|---|---|---|
| -8 | 0.451 | 0.164 |
| -6 | 0.349 | 0.134 |
| 0 | 0.323 | 0.101 |
| 6 | 0.279 | 0.098 |
| 12 | 0.227 | 0.099 |

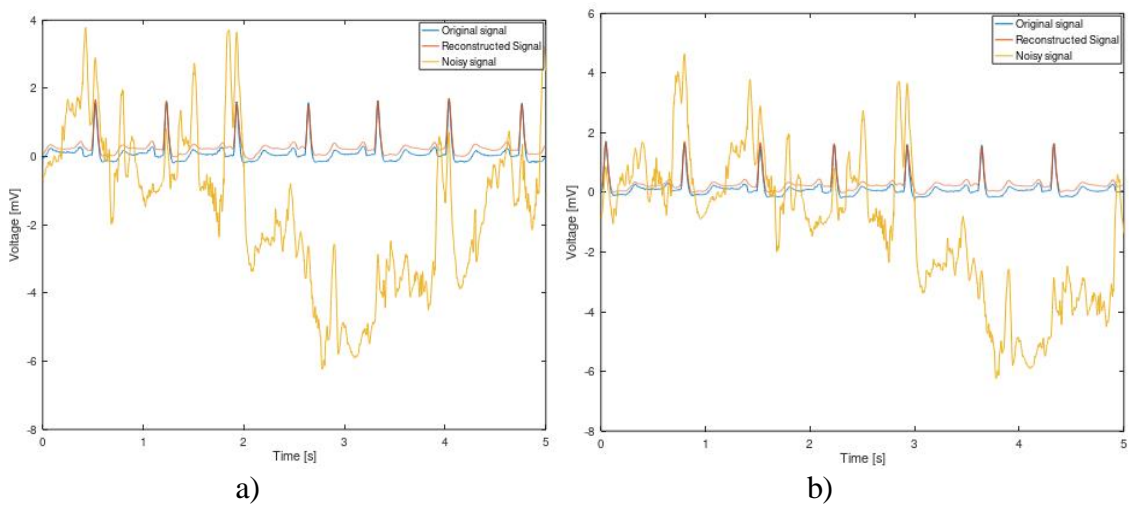

a)  b)

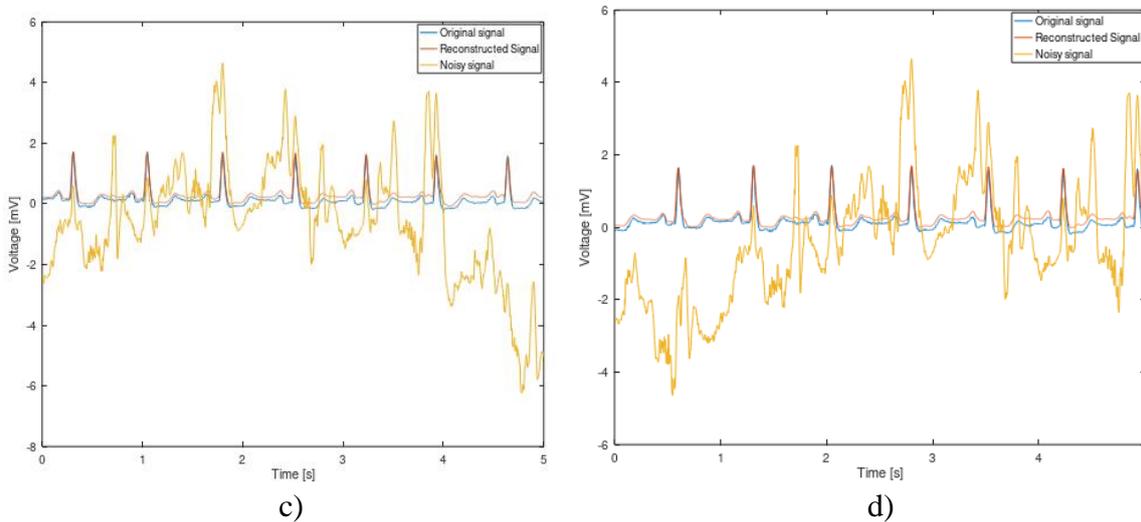

c)                                          d)

Figure 18. Comparison between the denoised/reconstructed ECG signal, the original/clean ECG signal and the noisy ECG signal for SNR = -6 and for record 105 of MIT PhysioNet database: a) RMS=0.1799; b) RMS = 0.1650; c) RMS =0.15506; d) RMS =0.14035.

## *4.4 Train and test on 10 records of MIT PhysioNet database and test on 1 record*

It was noticed in the previous sections that it is possible to train on an ECG record and test successfully on the same record with very good RMS and SNR values. However, it is of interest also to investigate whether it is possible to train on multiple ECG records and test on the same records and moreover if it is possible to train on multiple ECG records and test on another different ECG record and what are the RMSs and the SNRs values.

Ten ECG records from MIT PhysioNet database are used namely 103, 105, 111, 116, 122, 205, 213, 219, 223 and 230.

The noise characteristic of the electrode movement artefact was added to the clean records of each of the above records with 14 levels of SNRs: 36, 24, 20, 18, 14, 12, 8, 6, 3, 0, -1, -3, -6, -8.

This resulted in a large dataset of 100800 ECG signals with duration of 1 second and 360 samples per second. This large dataset was divided 3/4 in training dataset (i.e. 75600 features) and 1/4 in testing dataset (i.e. 25200 features).

The CNN model used has 3 convolutional layers each with 36 filters, size of kernel is [23 23] per each filter. The full description of the CNN model is shown in Table 9. The model was trained for 250 epochs (i.e. 94400 iterations) which took 1 hour and 53 minutes to reach an RMS equals 4.39 for the mini-batch data and loss equals 8.5.

TABLE 9. LISTING OF THE CNN LAYERS: $M = 360$ IS THE NUMBER OF SAMPLES PER INPUT ECG SIGNAL WITH DURATION OF 1 SECOND

| Nr | Type | Description |
|---|---|---|
| 1 | Image Input | 360x1x1 images with 'zerocenter' normalization |
| 2 | Convolution | 36 23x23x1 convolutions with stride [1 1] and padding 'same' |
| 3 | Batch Normalization | Batch normalization with 36 channels |
| 4 | ReLu | Rectified Linear Unit |
| 5 | Average Pooling | 1x1 average pooling with stride [4 1] and padding [0 0 0 0] |
| 6 | Convolution | 36 23x23x1 convolutions with stride [1 1] and padding 'same' |
| 7 | Batch Normalization | Batch normalization with 36 channels |
| 8 | ReLu | Rectified Linear Unit |
| 9 | Average Pooling | 1x1 average pooling with stride [4 1] and padding [0 0 0 0] |
| 10 | Convolution | 36 23x23x1 convolutions with stride [1 1] and padding 'same' |
| 11 | Batch Normalization | Batch normalization with 36 channels |
| 12 | ReLu | Rectified Linear Unit |
| 13 | Average Pooling | 1x1 average pooling with stride [4 1] and padding [0 0 0 0] |
| 14 | Fully Connected | 360 fully connected layer |
| 15 | Regression Output | Mean-squared-error with response 'Response' |

In Fig. 19 there are shown several comparisons between the denoised ECG signals, the noisy ECG signals and the clean ECG signal and with the associated RMS/SNR values: first figure RMS=0.114, SNR=11.41, second figure RMS=0.100, SNR=10.50, third figure RMS = 0.280, SNR=11.80 and the last figure RMS=0.280, SNR=11.80.

The average RMS over the testing dataset is 0.2112 and the average SNR over the same testing dataset is 7.02.

These values, as expected, are worse than the average RMS and SNR values obtained for single records and shown in the previous sections.

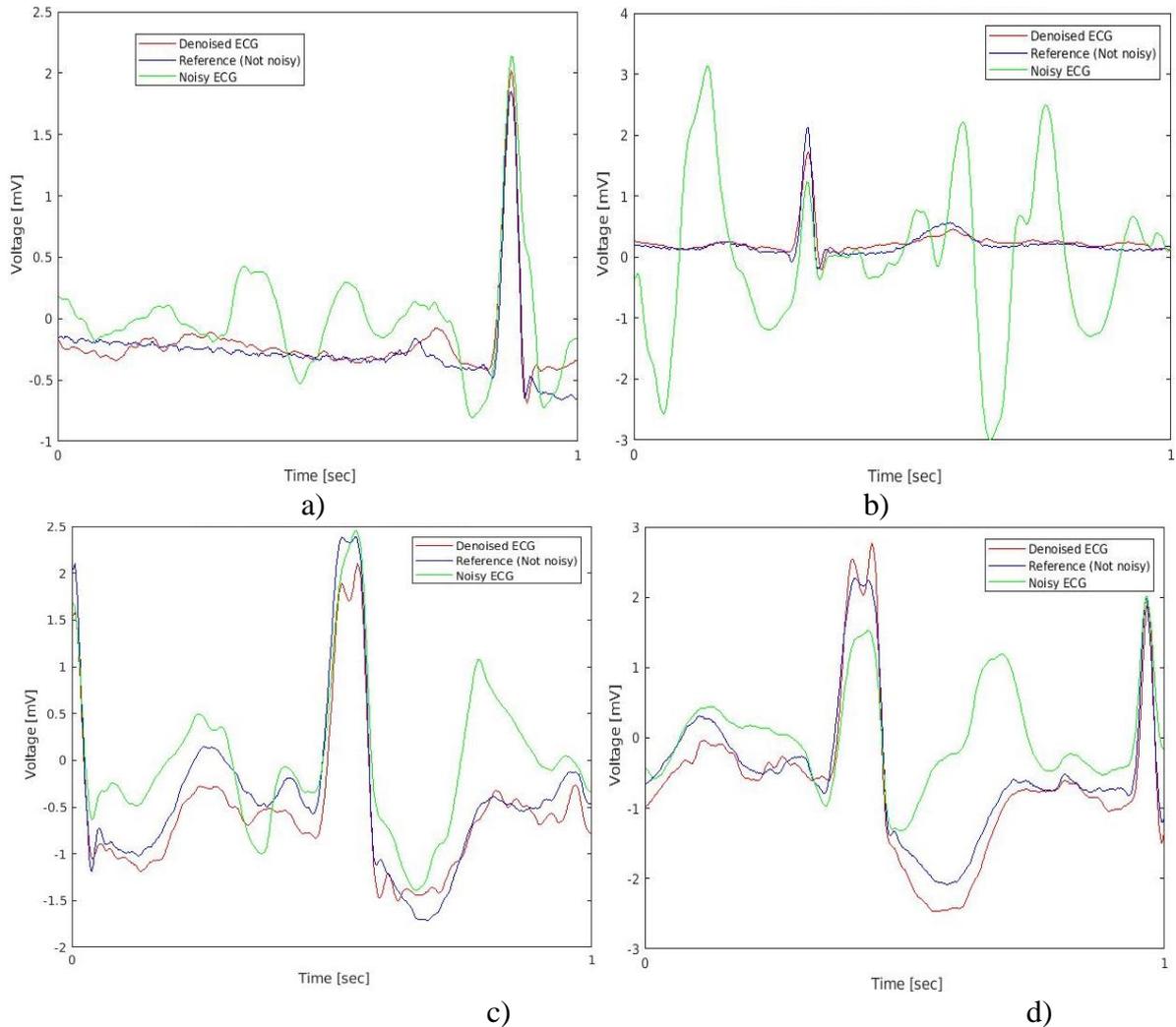

Figure 19. Training on 10 records of MIT PhysioNet database and testing on the same records: a) RMS = 0.114, SNR=11.41; b) RMS = 0.100, SNR=10.50; c) RMS = 0.277, SNR=10.91; d) RMS = 0.280, SNR=11.80.

In the following simulation, it was of interest to train the CNN model on the full input dataset of 100800 input features and test on a different record of MIT PhysioNet database. The CNN model was left to train again for 250 iterations (i.e. 94400 iterations) which took about 2 hours to reach an RMS of 4.1 for the mini-batch data and loss function equals 8. Testing on a dataset of the size of the previous training dataset that is size [25200x360] (i.e. the testing dataset was also included in the initial full training data) resulted in an averaged testing RMS = 0.1815 and average SNR = 8.1939 which values are obviously slightly better than in the first simulation (i.e. average RMS = 0.2112, average SNR = 7.02) as the training was done on the entire available dataset and the testing was done on a sample taken from this training dataset.

Finally, it was of interest to test on a different record, which in this case is taken as being record 100 of MIT PhysioNet database and for the level of SNR= -6 in the ECG signals of the respective record.

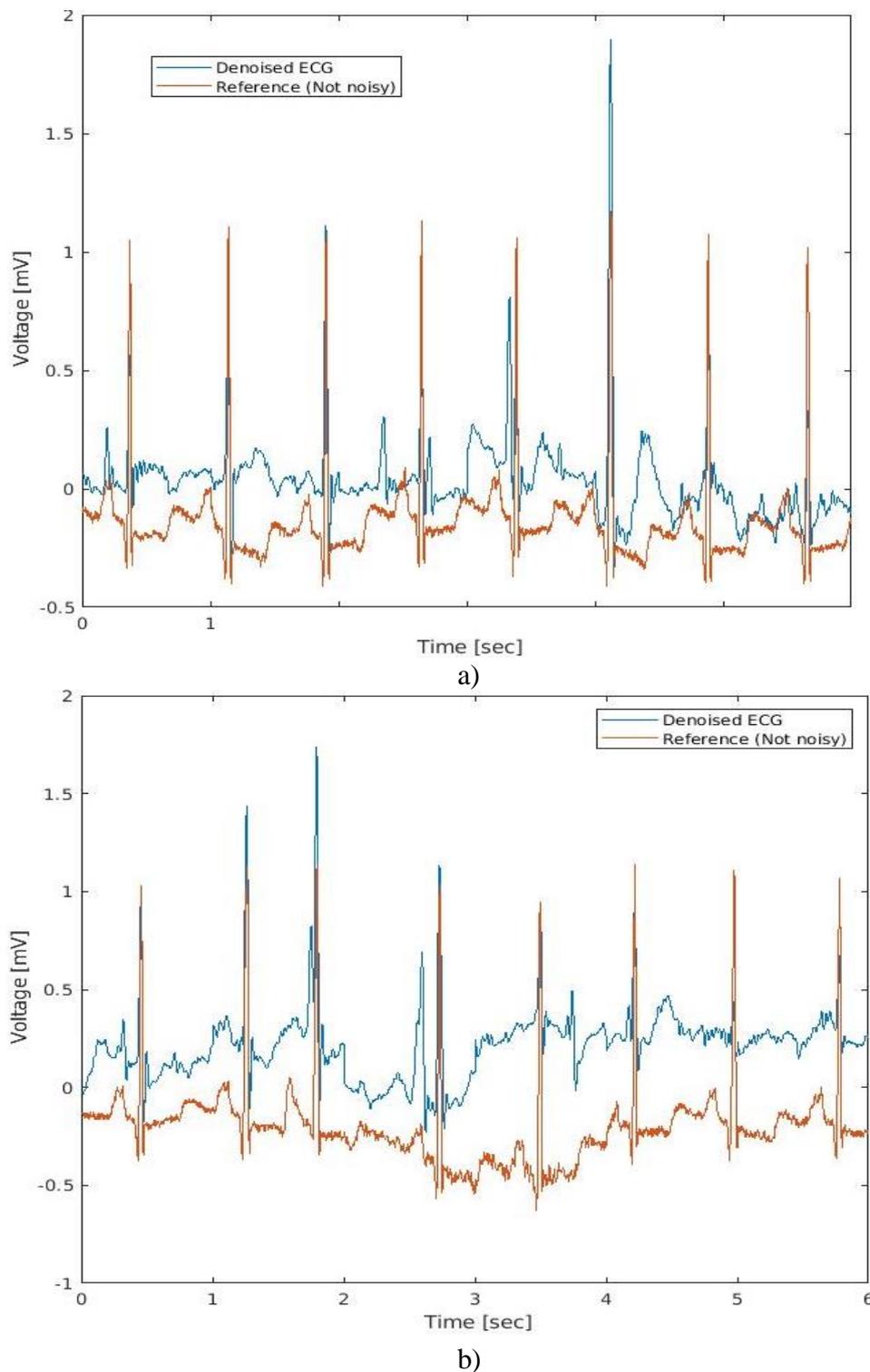

Figure 20. Testing on record 100 of MIT PhysioNet database: a) RMS = 0.231, SNR = -1.81; b) RMS = 0.4575 , SNR= -3.85.

In Fig. 20, it is shown a particular result of the testing on record 100 of MIT PhysioNet database: for the result in Fig. 20(a) there is RMS = 0.231 and SNR = -1.81 and in Fig. 20(b) there is RMS = 0.4575 and SNR= -3.85. Although it is possible to identify most of the spikes in the ECG signals in both figures, it can be observed the high differences between the clean and the denoised signals as well as the low values for the SNRs. In fact over the entire record 100 the average RMS was 0.3360 and the average SNR was -2.0776. It becomes clear that as there could be millions of ECG signals with different characteristics in the real world because of the multitude of different people (i.e. age, weight, pathological condition) being investigated or because even of the different ECGs machines being used, it will not be an easy task to approximate all the ECG signals by training a single DL model. Therefore in the following section it is proposed to train a DL model on a patient while at rest and then to use the respective ECG signal and to generate multiple ECG signals with various heartbeats and levels of noise, which would characterize the same patient in different conditions (i.e. running, waking, etc). This would form a rather synthetic ECG dataset but similar to the one from real life, and with which would be possible to train a DL model. Finally, the trained DL model will be used on the same patient when the patient is at effort for the purpose of filtering any noise, which can appear while the patient is doing an effort.

## *4.5   Denoising of multiple heartbeat ECG signals*

For real-time applications, the task of denoising ECG signals becomes more cumbersome, as the clean part of an ECG signal is very probable to be very limited in size so that to train on it. Therefore a solution put forth in this work is to train on 1 second ECG noisy artificial multiple heartbeat data (i.e. ECG at effort), which was generated in a first instance based on few sequences of real ECG data (i.e. ECG at rest). This corresponds also to reality, where usually the ECG human is put at rest and the ECG is recorded and then the same human is asked to do some exercises and the ECG is recorded at effort.

In Fig. 21(a), clean and noisy ECG signals are shown, with duration of 10 seconds from record 118 of MIT PhysioNet database: in Fig. 21(a) there are shown 12 heartbeats per 10 seconds, in Fig. 21(b) there are shown 13 heartbeats per 10 seconds while in Fig. 21(c) there are shown 15 heartbeats per 10 seconds. The SNR value for the noisy ECG signal was -6 in all three figures.

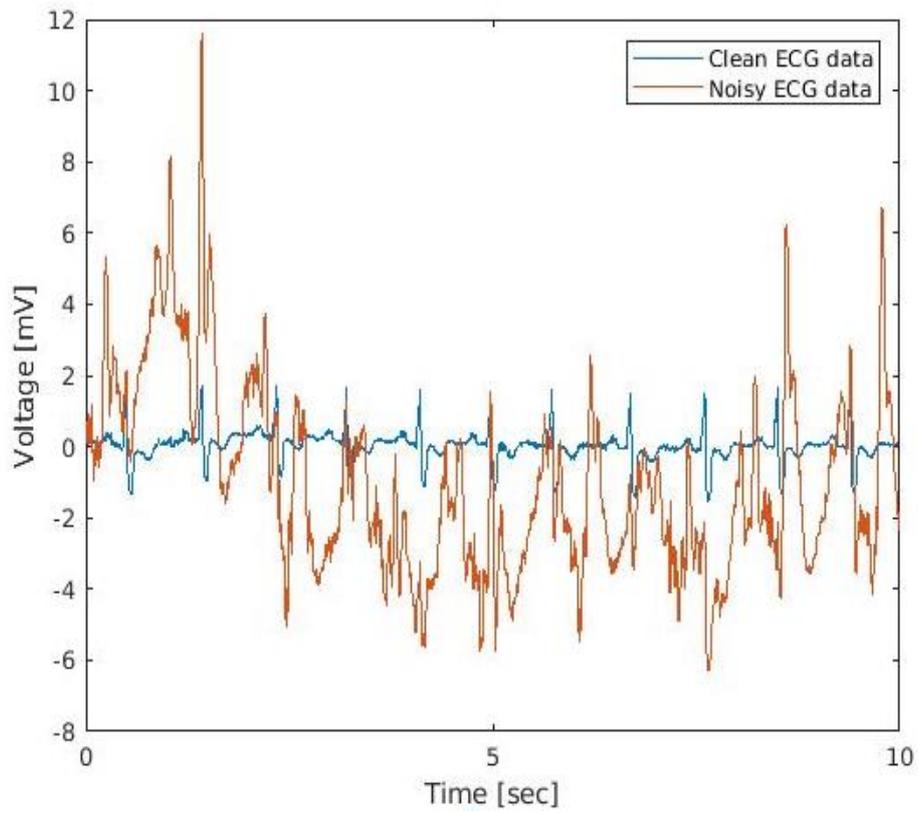

a)

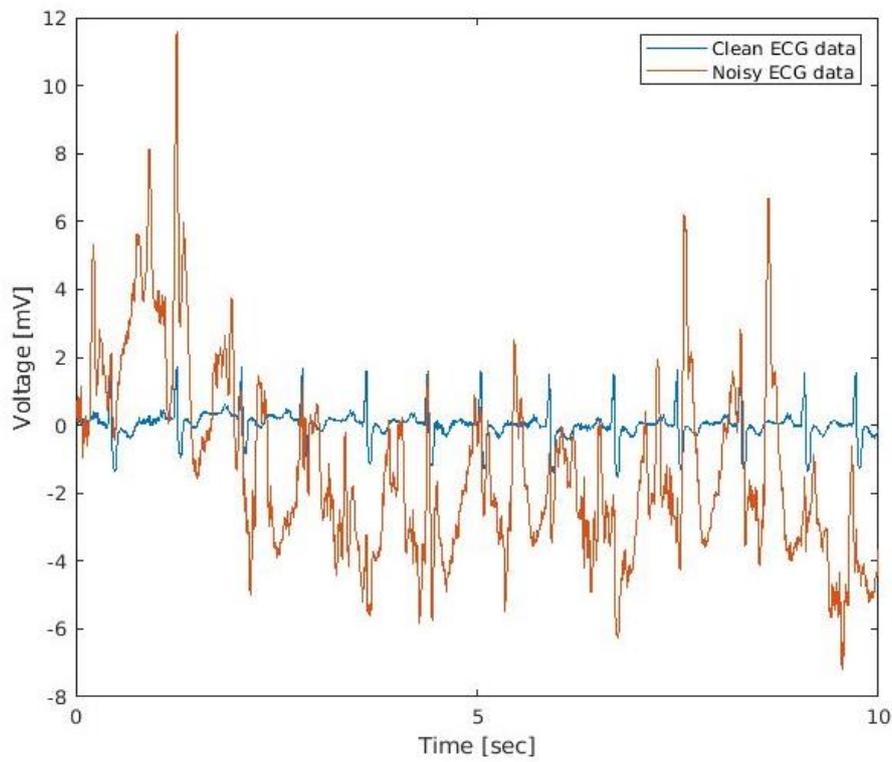

b)

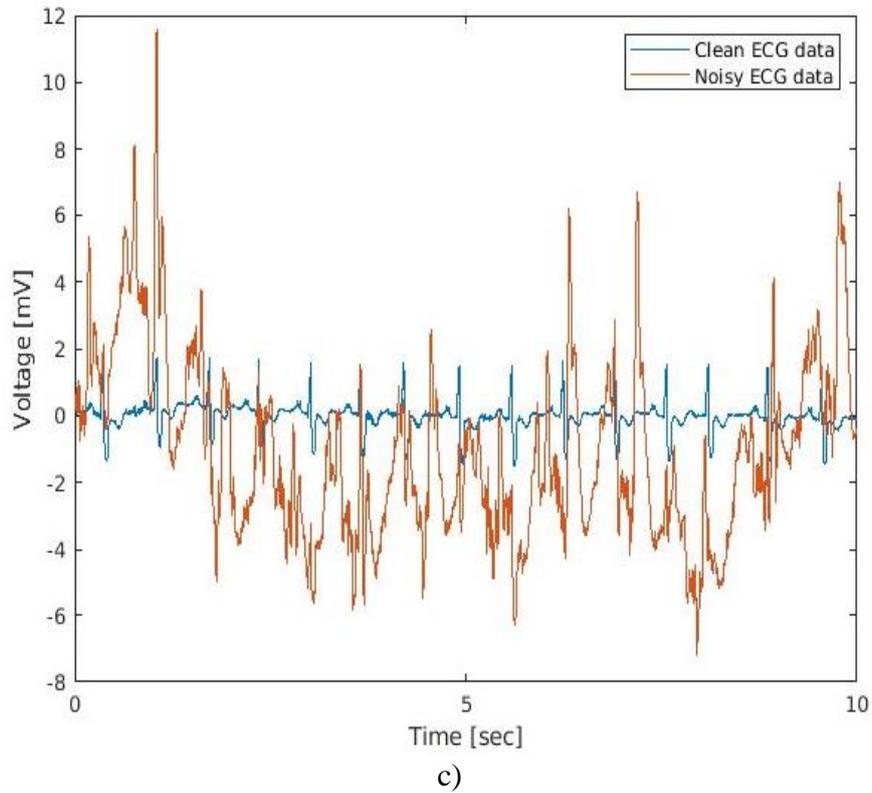

c)

Figure 21. Clean and noisy ECG signal with duration of 10 seconds from record 118 of MIT PhysioNet database: a) 12 heartbeats per 10 seconds; b) 13 heartbeats per 10 seconds; c) 15 heartbeats per 10 seconds.

The number of heartbeats was varied between 12 heartbeats per 10 seconds to 15 heartbeats per 10 seconds that is 72 heartbeats per minute to 90 heartbeats per minute. There are used 4 levels of SNRs (-6, 0, 6, 12) for the noisy data. This results in a dataset of 14256 clean different training ECG signals (i.e. and another 14256 noisy ECG signals), each ECG signal with a duration of 1 second and consisting of 360 sample points. This dataset (I) is divided into four parts: three parts of size 10692 representing the training dataset and the remaining 3564 signals representing the testing dataset.

The structure of the CNN model used to train is similar as to the previous ones: 3 CNN layers, each layer with 36 filters, each filter with [23 23] kernels. The convolutional layer has the same as before stride [1 1] and padding 'same', the average pooling layer is of size [1 1] and stride [4 1] and padding [0 0 0 0]. The regression output layer calculates the mean-squared-error for response. The batch size during the training is 200. Other parameters used during training are gradient threshold of 1 and initial learn rate of 0.01 and Adam optimization algorithm.

The RMS is calculated between each clean ECG signal and the predicted denoised ECG signal from the testing dataset. After 2000 epochs the RMS calculated for the mini-batch data by the CNN is 2.07 and the loss function becomes 2.1 . The 2000 epochs took 2 hours to accomplish. The average RMS over the testing dataset is 0.1781 and average SNR over the entire testing dataset is 8.2.

The average RMS over the entire testing dataset is 0.1761 after 20000 epochs which took about 23 hours to run. Therefore even with 2000 epochs and 2 hours is possible to get similar results in terms of quality.

In Fig. 22 is shown clean, denoised and noisy ECG signal with duration of 1 second from the record 118 of MIT PhysioNet database for multiple heartbeat ECG signal varying from 12 heartbeats per 10 seconds to 15 heartbeats per 10 seconds with the associated RMS and SNR values. Furthermore, for illustrative purposes, in Fig. 23 is shown a sequence of 5 seconds denoised ECG signal for multiple heartbeat ECG signal produced artificially and based on record 118 of MIT PhysioNet database. The denoised ECG signal obtained with the CNN model has an average RMS over the 5 seconds of 0.1221 and the average SNR over the 5 seconds of 10.49 between the denoised and clean ECG signal (Fig.22(a)). In Fig. 22(b) the RMS and SNR between the clean and denoised ECG signals are 0.1532 and 8.54.

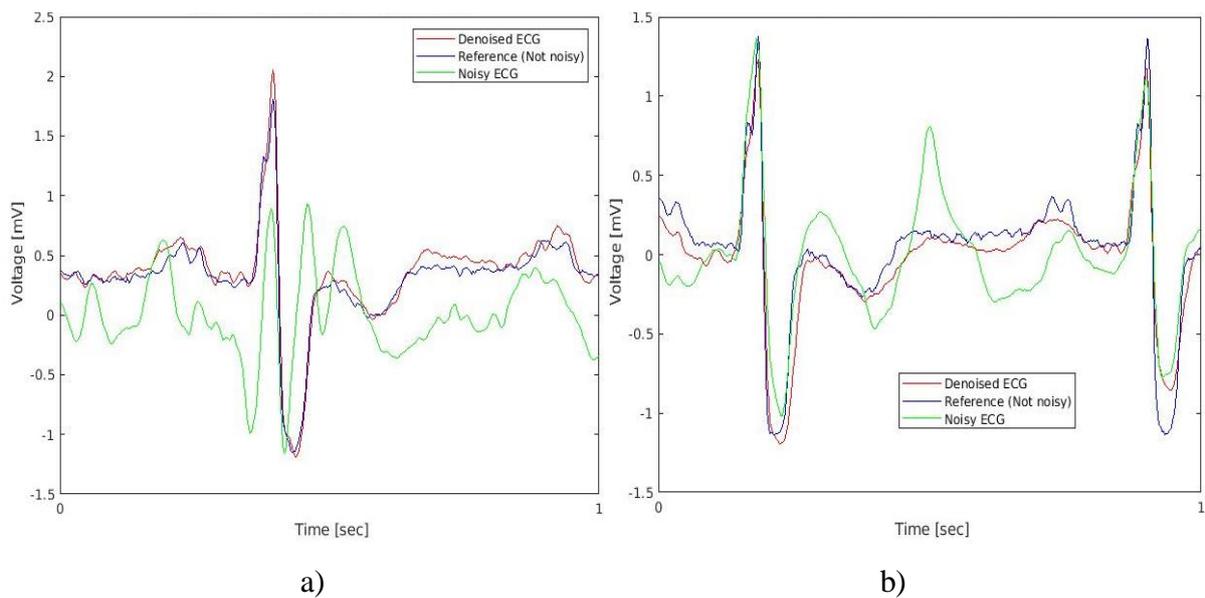

a)                                      b)

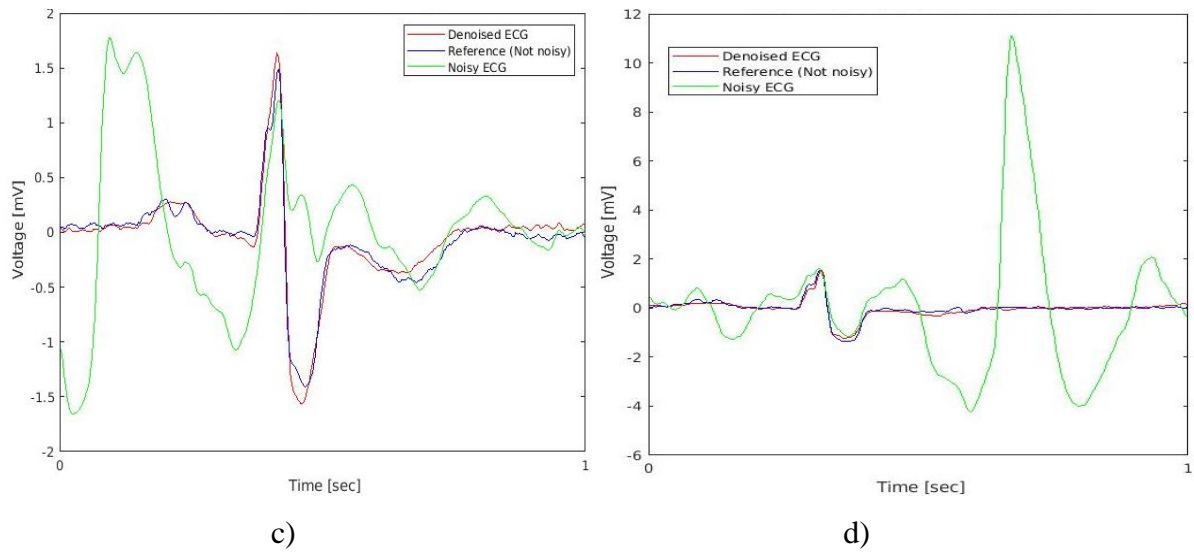

| c) | d) |

Figure 22. Clean, denoised and noisy ECG signals with duration of 1 seconds from record 118 of MIT PhysioNet database: a) RMS = 0.0852 , SNR= 15.12; b) RMS = 0.1316, SNR =10.20; c) RMS =0.0855 , SNR =13.72, d) RMS = 0.0847, SNR = 13.22.

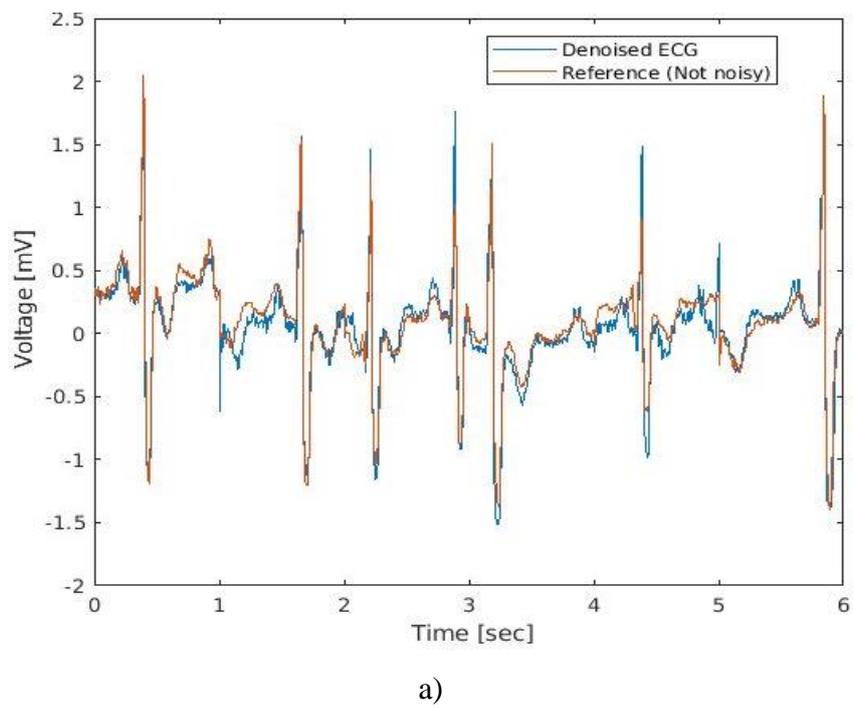

a)

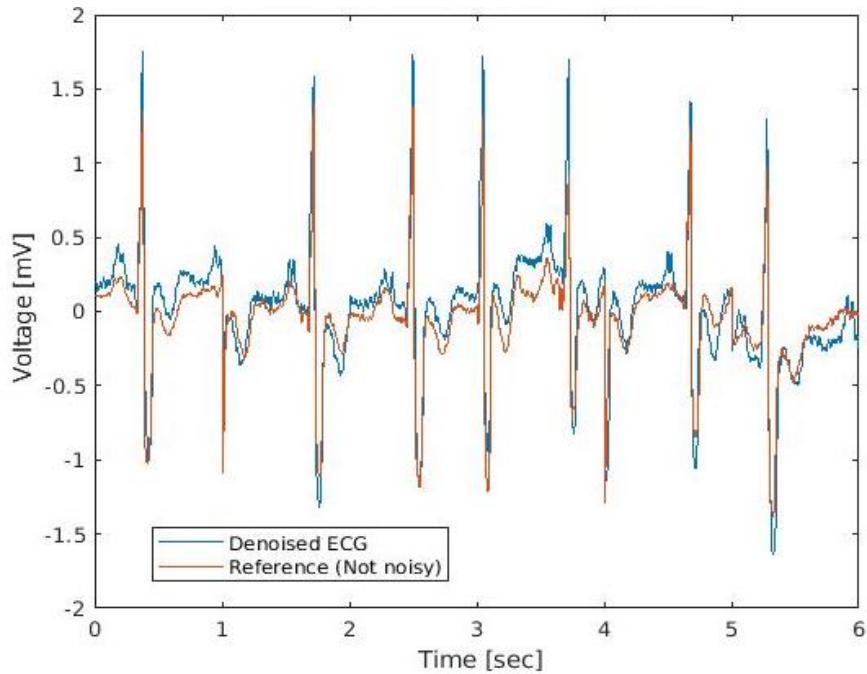

b)

Figure 23. Sequence of 5 seconds denoised ECG signals with CNN model for multiple heartbeat ECG signal produced artificially and based on record 118 of MIT PhysioNet database: a) RMS = 0.1221, SNR= 10.49; b) RMS = 0.1532, SNR = 8.54.

As it was noticed before (i.e. train/test on 10 records of MIT PhysioNet database and test on 1 record), because there could be a limited number of noisy ECG signals to train on, then the CNN model finds it difficult to generalize well to all the testing signals. The solution to this situation is to generate more noisy patterns of data so that to train the CNN model on this extended set of data and by doing so to improve the accuracy of the results of the CNN model.

Therefore a new multiple heartbeat dataset was generated based on the above dataset by considering more noisy ECG signals obtained from the clean ECG signals to which there was added the electrode motion artefact noise and also some random noise. This resulted in a new training/testing dataset of 42768 ECG signals (i.e. duration 1 second) from which 32076 ECG signals represented the training dataset and 10692 ECG signals represented the testing dataset. The CNN model was left to train for 322 epoch (i.e. 51000 iterations) which took 4 hours and 52 minutes to run so that to reach an RMSE of 2.43 and loss function of 3 as calculated by the Matlab DL algorithm for the mini-batch dataset. The average RMS over the testing dataset was 0.1419 and the average SNR over the testing dataset was SNR = 9.6706 dB values. These values, are slightly better than the ones obtained for the first dataset from above, which consisted of only 10692 ECG training signals and 3564 testing ECG signals and for which the average RMS over the testing dataset was 0.1781 and the average SNR over the entire testing dataset was 8.2. Therefore, with more input features to be trained on the CNN model, it is expected that the accuracy of results to increase.

Finally, for the same dataset with 42768 ECG signals from which 32076 ECG signals represented the training dataset and 10692 ECG signals represented the testing dataset, after 200 epochs (i.e. 33250 iterations), which took 1 hour and 41 minutes to complete, the RMS as calculated by the Matlab DL toolbox was 2.38 and the loss function was 2.8. These values are similar to the ones reported above for the same dataset after 322 epoch (i.e. 51000 iterations), which took 4 hours and 52 minutes. This clearly suggests that using other alternative measures of performance such as RMS and SNR over the testing datasets can represent complementary and adequate ways of measuring the accuracy of the results that are the denoised ECG signals.

## *5. Conclusions*

Denoising of ECG data is a paramount task for the purposes of correct interpretation and classification of the ECG status of a person and regardless of the medical condition of the respective person.

In this paper an evaluation of two DL models has been implemented for denoising of ECG data: CNN and LSTM. The CNN model results on both synthetic and real data were compared to the LSTM model results and it showed that the CNN model produced better predictions or at least as good as the other one. A comparison with an established technique for denoising of ECG data, namely Wavelet denoising method, proved that while for a specific level of random noise, the Wavelet method is a suitable technique, then for higher levels of random and drifting noise the DL models may represent a definite advantage.

It was shown that it is possible to train a CNN model on 1 second ECG noisy artificial multiple heartbeat data (i.e. ECG at effort), which was generated in a first instance based on few sequences of real ECG data (i.e. ECG at rest). Then, it was possible to use the trained CNN model to denoise the ECG signal, which had also multiple heartbeats. This situation would be, for example, very useful for denoising in real-time ECG data coming from a Holter dispositive [39]. Moreover, it could be also used for a person who is doing physical exercises and where initially the person would have recorded few clean ECG sequences, which would be used to generate a series of artificial multiple heartbeat ECG sequences. With the artificial ECG data, a CNN model would be trained. In turn, the trained CNN model would be used to denoise ECG data coming from the same person who would perform some physical exercises (i.e. ECG recorded at effort) and the noise will be rejected in real-time.

Further work, it will involve training/testing on other ECG records and with different SNR values and types of errors. Moreover, in [17] the ECG signals were scaled between 0 and 1 and very good results were reported for the denoised ECG signals: therefore, it is paramount to investigate some of the previous presented results not only by introducing a scaling of the ECG signals such as between 0 and 1 but to search for other possible benefits from using other pre-filtering techniques then the ones used and introduced in Algorithm 1.

The computational times were drastically reduced by using an advanced GPU (Graphical Processing Unit) that is an NVIDIA TITAN V GPU with 12 GB RAM, although some of the simulations were left to run for longer times so that to investigate the RMS and the loss error function of the mini-batch data, and especially the accuracy of numerical results achieved after longer runs.

Finally, as an element of novelty, the paper presented also a Design of Experiment (DoE) study which was able to determine the optimal structure of a CNN model with regard to the accuracy of results and the computational times and for the considered problem of denoising the real ECG signals, and which type of study has not been seen in the literature before for DL model.

Acknowledgment

The author would like to thank to the Engineering and Physical Sciences Research Council (EPSRC) Impact Acceleration Account (IAA 220) from the University of Manchester for supporting this work.